\documentclass{article}

\usepackage{microtype}
\usepackage{graphicx}
\usepackage{subfigure}
\usepackage{booktabs} 

\usepackage{hyperref}

\usepackage[accepted]{icml2024}

\usepackage{amsmath}
\usepackage{amssymb}
\usepackage{mathtools}
\usepackage{amsthm}
\usepackage{xspace}

\usepackage[capitalize,noabbrev]{cleveref}
\newcommand{\hieros}{\textsc{Hieros}\xspace}
\newcommand{\repourl}{\url{https://github.com/Snagnar/Hieros}}

\theoremstyle{plain}

\theoremstyle{definition}

\theoremstyle{remark}

\icmltitlerunning{Hieros: Hierarchical Imagination on Structured State Space Sequence World Models}

\begin{document}

\twocolumn[
\icmltitle{Hieros: Hierarchical Imagination on Structured State Space Sequence World Models}




\begin{icmlauthorlist}
\icmlauthor{Paul Mattes}{yyy}
\icmlauthor{Rainer Schlosser}{yyy}
\icmlauthor{Ralf Herbrich}{yyy}
\end{icmlauthorlist}

\icmlaffiliation{yyy}{Digital Engineering Faculty, Hasso Plattner Institute, University of Potsdam, Germany}

\icmlcorrespondingauthor{Paul Mattes}{paul.mattes@student.hpi.de}
\icmlcorrespondingauthor{Rainer Schlosser}{rainer.schlosser@hpi.de}

\icmlkeywords{Machine Learning, ICML, Reinforcement Learning, World Models, Structured State Space Sequence Models}

\vskip 0.3in
]



\printAffiliationsAndNotice{\icmlEqualContribution} 

\begin{abstract}
    One of the biggest challenges to modern deep reinforcement learning (DRL) algorithms is sample efficiency. Many approaches learn a world model in order to train an agent entirely in imagination, eliminating the need for direct environment interaction during training. However, these methods often suffer from either a lack of imagination accuracy, exploration capabilities, or runtime efficiency. We propose \hieros, a hierarchical policy that learns time abstracted world representations and imagines trajectories at multiple time scales in latent space. \hieros uses an S5 layer-based world model, which predicts next world states in parallel during training and iteratively during environment interaction. Due to the special properties of S5 layers, our method can train in parallel and predict next world states iteratively during imagination. This allows for more efficient training than RNN-based world models and more efficient imagination than Transformer-based world models. 
    We show that our approach outperforms the state of the art in terms of mean and median normalized human score on the Atari 100k benchmark, and that our proposed world model is able to predict complex dynamics very accurately. We also show that \hieros displays superior exploration capabilities compared to existing approaches.
 \end{abstract}

\begin{figure*}[ht!]
    \centering
        \includegraphics[width=0.52\textwidth]{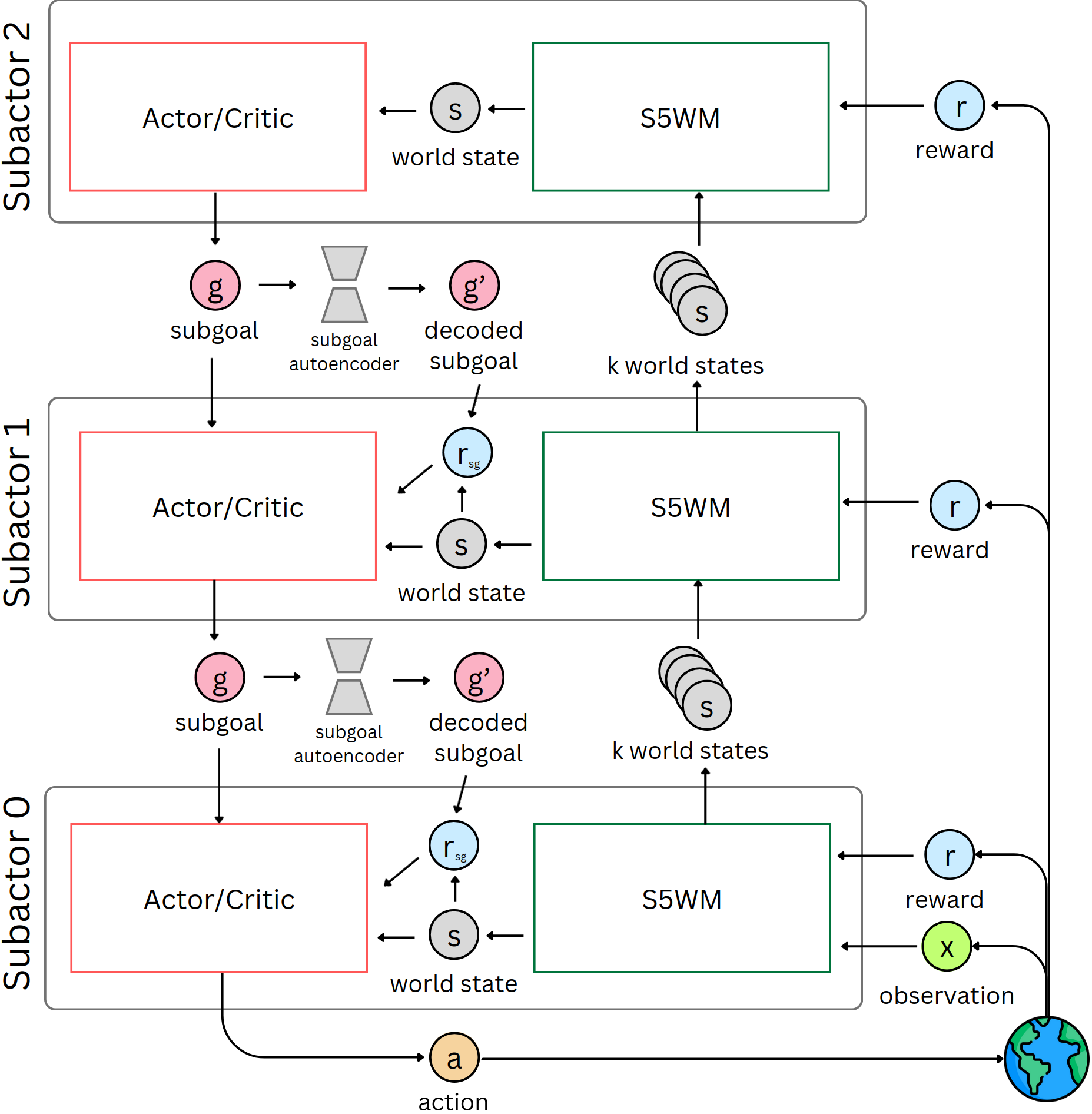}
        \hspace{0.3cm}
        \includegraphics[width=0.35\textwidth]{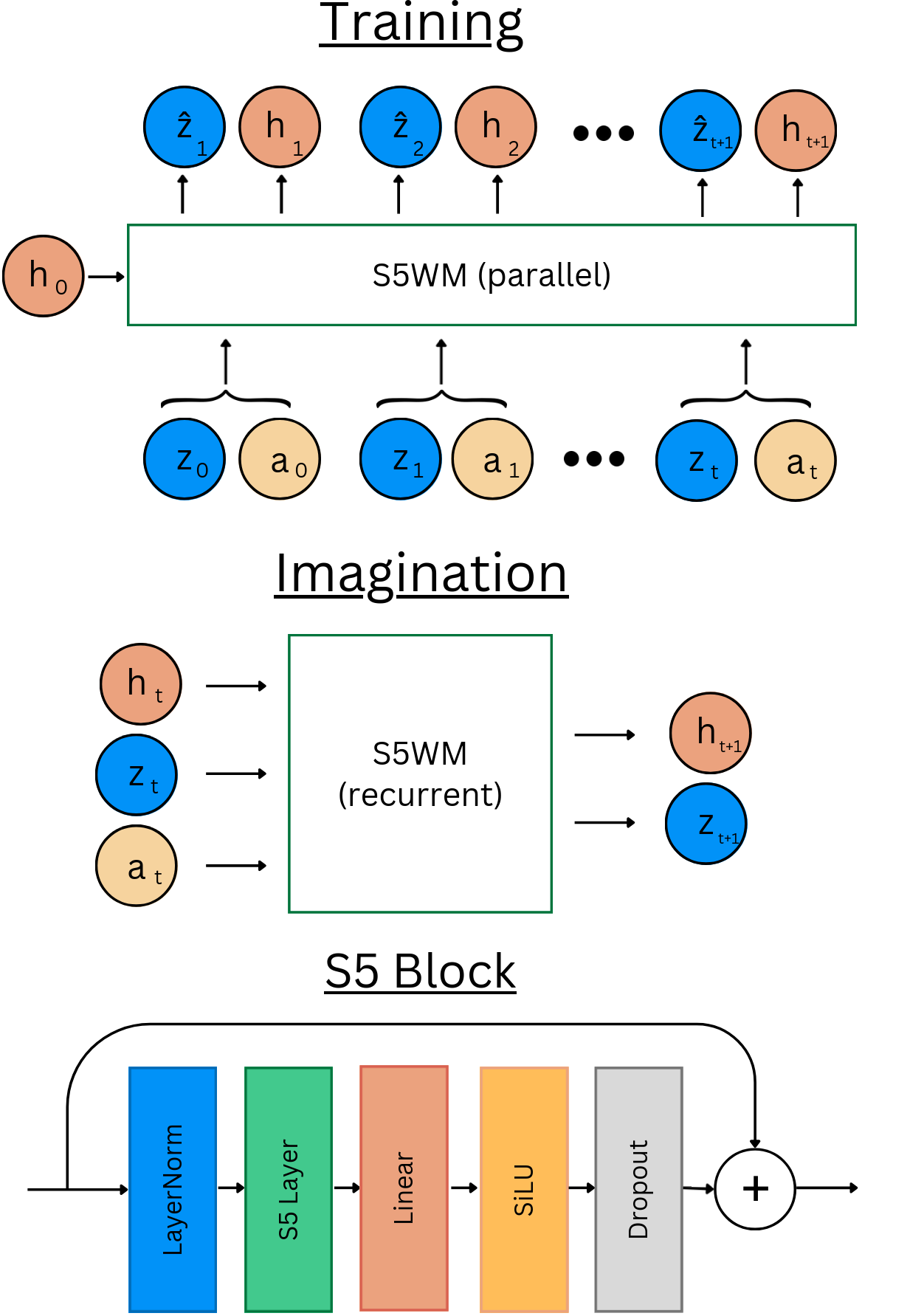}
\vspace{-0.1cm}
        \caption{On the left: Hierarchical subactor structure of \hieros. Each layer of the hierarchy learns its own latent state world model and interacts with the other layers via subgoal proposal. The action outputs of each actor/critic is the subgoal input of the next lower layer. The output of the lowest level actor/critic is the actual action in the real environment. On the right: Training and imagination procedure of the S5WM. \hieros uses a stack of S5 blocks with their architecture shown above. 
        \vspace{-0.3cm}}
\label{fig:hieros_structure}   
\end{figure*}

\section{Introduction}
\label{sec:introduction}

Learning behavior from raw sensory input is a challenging task. Reinforcement learning (RL) is a field of machine learning that aims to solve this problem by learning a policy that maximizes the expected cumulative reward in an environment \citep{sutton2018reinforcement}. The agent interacts with the environment by taking actions and receiving observations and rewards. The goal of the agent is to learn a policy that maximizes the expected cumulative reward. The agent can learn this policy by interacting with the environment and observing 
rewards, which is called model-free RL. 
Some agents also learn a model of their environment and learn a policy based on simulated environment states. This is called model-based 
RL 
\citep{moerland2023model}. 

However, a significant hurdle encountered by RL algorithms is the lack of sample efficiency \citep{micheli_transformers_2023}. The demand for extensive interactions with the environment to learn an effective policy can be prohibitive in many real-world applications \citep{yampolskiy2018artificial}. In response to this challenge, deep reinforcement learning (DRL) 
has emerged as a promising solution. DRL leverages neural networks to represent and approximate complex policies and value functions, allowing it to tackle a wide array of problems and environments effectively.

One such approach is the concept of ``world models''. World models aim to create a simulated environment within which the agent can generate an infinite amount of training data, thereby reducing the need for extensive interactions with the real environment \citep{ha2018world}. However, a key prerequisite for world models is the construction of precise models of the environment, a topic that has garnered substantial research attention \citep{hafner_learning_nodate,hafner_dream_2020,kaiser2019model}. The idea of learning models of the agent's environment has been around for a long time \citep{nguyen1990truck, schmidhuber1990line, forward}. After being popularized by \citet{ha2018world}, world models have since evolved and diversified to address the sample efficiency problem more effectively. Most prominently, the DreamerV1-3 models \citep{hafner_dream_2020,hafner_mastering_2022,hafner_mastering_2023} have achieved state-of-the-art results in multiple benchmarks such as Atari100k \citep{bellemare2013arcade} or Minecraft \citep{kanitscheider2021multi}. These models use a recurrent state space model (RSSM) \citep{hafner_learning_nodate} to learn a latent representation of the environment. The agent then uses this latent representation to train in imagination. Recently, Transformers have gained popularity as backbones for world models, due to their ability to capture complex dependencies in data. We will discuss some approaches using Transformer-based architectures in \Cref{sec:related_work}. One of the major drawbacks of those architectures is the inherent lack of runtime efficiency of the attention mechanism used in Transformers. Recently proposed structured state space sequence (S4) models \citep{gu_efficiently_2022} show comparable or superior performance in a wide range of tasks while being more runtime efficient than Transformer-based models, which makes them a promising alternative 
\citep{deng_facing_2023,lu_structured_2023}.

Another avenue for improving RL sample efficiency is hierarchical RL (HRL) \citep{dayan1992feudal,parr1997reinforcement, sutton1999between}. This approach operates at different time scales, allowing the agent to learn and make decisions across multiple levels of abstraction. The idea is that higher level policies divide the environment task into smaller subtasks or subgoals (also commonly called skills). The lower level policy is then rewarded for fulfilling these subgoals and is thus guided to fulfill the overall environment task. This approach has been shown to be effective in a variety of tasks \citep{hafner_deep_2022,nachum_data-efficient_2018,jiang_language_2019,nachum_multi-agent_2019}. \citet{hafner_deep_2022} proposes an HRL approach that builds on DreamerV2 \citep{hafner_mastering_2022} and achieves superior results in the Atari100k benchmark.

Finding the right experience replay buffer sampling strategy is key for many RL algorithms, as it has a great influence on the final performance of the agent \citep{fedus2020revisiting, d2022sample}. \citet{li_transformers_2023} introduce a time balanced replay dataset which empirically boosted the performance of their imagination based RL agent. However, this replay procedure relies on recomputing all probabilities in $O(n)$ at each iteration, which reduces its applicability for other approaches.

\citet{lecun2022path} theorizes that an HRL agent that learns a hierarchy of world models and features intrinsic motivation to guide exploration could potentially achieve human-level performance in a wide range of tasks. \citet{nachum_why_2019} and \citet{aubret_information-theoretic_2023} provide further reasoning that combining HRL with other successful approaches such as world models could lead to a significant improvement in sample efficiency.
Motivated by these findings, we propose \hieros, a multilevel HRL agent that learns a hierarchy of world models, which use S5 layers to predict next world states.
Specifically, our contributions are as follows:
\begin{itemize}
    \setlength{\itemsep}{1pt}  %
	\setlength{\parskip}{0pt}
	\setlength{\parsep}{0pt}
    \item We introduce \hieros, a Hierarchical Reinforcement Learning (HRL) agent designed to learn a hierarchy of world models, facilitating the acquisition of complex and temporally abstract behaviors. This enables \hieros to project future environment states both far into the future and accurately for near-term interactions. Notably, our architecture is the first of its kind to employ hierarchical imagination within a multilevel framework, characterized by more than two layers.
    \item Furthermore, we propose S5WM, a world model architecture that leverages S5 layers to forecast subsequent world states. This model exhibits several advantageous properties compared to the RSSM used in DreamerV3 \citep{hafner_mastering_2023} and several proposed Transformer-based alternatives \citep{micheli_transformers_2023,robine_transformer-based_2023,chen_transdreamer_2022} as well as a recently proposed S4WM \citep{deng_facing_2023}. Our novel contribution lies in leveraging the efficient design of our S5WM, capitalizing on its accessibility to the internal state, to effectively model environment dynamics in the context of imagination-based reinforcement learning.
    \item We derive efficient time-balanced sampling (ETBS) for experience dataset sampling from the time-balanced sampling method proposed by \citet{robine_transformer-based_2023} with a sampling time complexity of $O(1)$.
    \item We show that \hieros achieves a new state-of-the-art mean and median normalized human score in the Atari100k benchmark \citep{bellemare2013arcade}.
    \item We conduct a thorough ablation study that combining hierarchical imagination based learning and our S5WM yields superior results. We also test the impact of different design choices of \hieros (e.g., world model choice, hierarchy depth, sampling procedure).
\end{itemize}

We describe our method in detail in \Cref{sec:methodology}. 
In \Cref{sec:experiments}, we evaluate \hieros on the Atari100k benchmark \citep{bellemare2013arcade} and analyze our results.
Concluding remarks and ideas for future work are given in the final \Cref{sec:conclusion}. 
We provide
a section explaining some background concepts of the topic in \Cref{sec:background}.
In this introduction we already discussed several related works, however, we provide further comparisons with existing papers in \Cref{sec:related_work}.
In \Cref{sec:appC} we provide a wide range of ablation studies.

\section{Methodology}
\label{sec:methodology}

In the context of RL, an agent interacts with an environment at discrete time steps, denoted as $t$. For the Atari 100K benchmark, for instance, where the environment represents a game like Pong, the agent's interaction involves selecting an action $a$ at time $t$ within the game, similar to making in-game moves or pressing buttons. Subsequently, the agent receives an observation $o$ and a reward $r$ from the environment. In the case of Pong, $o$ typically takes the form of a pixel image capturing the game's visual state, while $r$ represents the points earned as a result of the agent's actions. The agent's primary goal is to learn an optimal policy $\pi$, guiding its interactions with the environment, with the overarching objective of maximizing the expected cumulative reward, expressed as $\mathbb{E}\left[\sum_{t=0}^\infty \gamma^t r_t\right]$, where $\gamma <1$ represents the discount factor and $r_t$ stands for the reward at time step $t$.

In this section we introduce HIERarchical imagination On Structured state space sequence world models (\hieros), a hierarchical model-based 
RL 
agent, that learns on trajectories generated by a Simplified Structured State Space Sequence (S5) model \citep{smith_simplified_2023}. 
We mainly base our approach upon DreamerV3 \citep{hafner_mastering_2023} and Director \citep{hafner_deep_2022}. In the following section, we propose two major changes to the DreamerV3 architecture. First, we describe a hierarchical policy, where each abstraction layer learns its own S5 world model and an actor-critic. Second, we replace the world model with a world model based on S5 layers. We also introduce efficient time-based sampling (ETBS) method for true uniform sampling over the experience dataset with $O(1)$ time complexity.

\subsection{Multilayered Hierarchical Imagination}
\label{sec:multilayered_hierarchical_imagination}

\hieros employs a goal conditioned hierarchical policy. Each abstraction layer learns its own S5 world model, an actor-critic and a subgoal autoencoder. We give some background on hierarchical reinforcement learning in \Cref{sec:background_hierarchical_reinforcement_learning}. The overall design of the subgoal proposal and the intrinsic reward computation is similar to the Director architecture \citep{hafner_learning_nodate}, which successfully implements a hierarchical policy on the basis of DreamerV2. In contrast to Director and many other works, our architecture can easily be scaled to multiple hierarchy levels. Most approaches apply only one lower level and one higher level policy \citep{hafner_deep_2022,nachum_data-efficient_2018,jiang_language_2019,nair_hierarchical_2019}.

In \Cref{fig:hieros_structure}, we illustrate the hierarchical structure and interaction among subactors. Each subactor $i$ comprises three modules: the world model ($w^i_\theta$), an actor-critic component ($\pi^i_\phi$), and a subgoal autoencoder ($g^i_\psi$). The world model imagines trajectories for the actor-critic training. The subgoal autoencoder compresses and decompresses states within the world model, with the decoder handling subgoal decoding, computing subgoal rewards $r_g$, and generating novelty rewards $r_{nov}$.
Only the lowest layer (subactor 0) interacts directly with the environment. The higher layers receive $k$ consecutive world model states from the lower level as observations, train their world model on these states and generate subgoals $g_i$. This is also in contrast to Director, which only provides the higher level policy with every $k$-th world state. We show the effect of providing all intermediate states as input for the higher level policy in \Cref{sec:appC:intermediate_encoding_vs_only_kth}.

The subgoals represent world states that the lower layer is tasked to achieve. They are kept constant for $k$ steps of the lower layer. So higher level subactors can only update their proposed subgoal every $k$ steps of the next lower level.
The actor-critic component is trained on imagined rollouts of the word model. The reward $r$ is a mixture of extrinsic rewards $r_{extr}$ (i.e. observed rewards directly from the environment), subgoal rewards $r_{g}$, and novelty rewards $r_{nov}$:
\begin{align}
    r = r_{extr} + w_{g} \cdot r_{g} + w_{nov} \cdot r_{nov}
\end{align}
Here, $w_{g}$ and $w_{nov}$ are hyperparameters that control the influence of the subgoal and novelty rewards on the overall reward.
The subgoal and novelty rewards $r_g$ and $r_{nov}$ are computed as follows:
\begin{align}
    r_{nov} = ||h_t - g^i_\psi(h_t)||_2 \ , \ r_g = \frac{g^T_t \cdot h_t}{\max(||g_t||, ||h_t||)}
\end{align}
with $h_t$ being the deterministic world model state, $g^i_\psi$ the subgoal autoencoder of subactor $i$, $g_t$ the subgoal and $||\cdot||_2$ being the L2 norm.
Since the subgoal autoencoder is trained to compress and decompress the world model state, it is able to model the distribution of the observed model states. This allows using its reconstruction error on the current world model state as a novelty reward $r_{nov}$.
The subgoal reward is computed using the cosine max similarity method between the world model state and the decompressed subgoal, which was proposed by \cite{hafner_deep_2022}.

The actor learns the policy:
\begin{align}
    a_t \sim \pi^i_\phi((h_t, z_t), g_t, r_t, c_t, H(p_\theta(z_t | m_t)))
\end{align}
with $h_t$ and $z_t$ representing the current model state, $g_t$ being the subgoal, $r_t$ being the reward, $c_t$ being the continue signal, and $H(p_\theta(z_t | m_t))$ being the entropy of the latent state distribution. The entropy term is used to encourage exploration and make the actor aware of states with high uncertainty. These additional inputs to the actor network are motivated by findings of \citet{robine_transformer-based_2023} which show, that providing the predicted reward as input improves the learned policy. The actor is trained using the REINFORCE algorithm \citep{williams1992simple}.
In the case of a higher level subactor, $a_t$ is the subgoal $g_t$ for the next lower layer. 
The actor-critic trains three separate value networks for each reward term 
to predict their future 
mean 
return.
The subgoal autoencoder $g^i_\psi$ is composed of an encoder and a decoder:
\begin{align}
        \text{Encoder: } g_t \sim p_\psi(g_t | h_t)\  , \ \text{Decoder: } h_t \approx f\psi(g_t)
\end{align}
The deterministic part $h_t$ of the model state is the only input for the subgoal autoencoder. This choice is made because the stochastic part $z_t$ is less controllable by the lower-level actor, making the subgoal less achievable.
\citet{hafner_deep_2022} found that the full latent state space of the lower level world model as action space would constitute a high dimensional continuous control problem for the higher level actor, which is hard for a policy to optimize on. Instead, the $g^i_\psi$ compresses the latent state into a discrete subgoal space of 8 categorical vectors of size 8. This compressed action space is much easier to navigate for the subgoal proposing policy than the continuous world state. While \citet{hafner_deep_2022} uses the decompressed goal as input for their worker policy, our approach with \hieros demonstrates improved subgoal completion and overall higher rewards when training subactors on the compressed subgoals. We directly compare these two approaches in \Cref{app:decompress}. The subgoal autoencoder is trained using a variational loss:
\begin{align}
    \mathcal{L}_{\psi} = ||\,f_\psi(z) - h_t\,||_2\, + \beta \cdot KL\left[\,p_\psi(\,g_t \,| \,h_t)\, || \,q(g_t)\,\right]
\end{align}
$z$ is sampled from the encoder distribution $z \sim p_\psi(g_t | h_t)$ and $q(g_t)$ is a uniform prior. $\beta$ is a hyperparameter that controls the influence of the KL divergence on the overall loss. All hyperparameters used in our experiments can be found in \Cref{sec:appB}.

The subgoals of all hierarchy layers can be decoded into the original image space of the game. This allows to visualize which subgoals the agent is trying to achieve, which makes the actions taken by \hieros explainable. We show some examples in \Cref{sec:app_visual}.

\subsection{S5-based World Model (S5WM)}
\label{sec:s5wm}

The world model in \hieros is responsible for imagining a trajectory of future world states. For this task, we use a similar architecture as DreamerV3 while swapping out the RNN-based sequence model for an S5-based sequence model. We give some background on world models and S4/S5 layers in \Cref{sec:background_world_models,sec:background_structured_state_space_sequence_models}. Our 
world model consists of the 
networks:
\begin{align*}
        \text{Sequence model: }\quad &(m_t, h_t)      = f_\theta(h_{t-1}, z_{t-1}, a_{t-1})\\
        \text{Encoder: }\quad &z_t                   \sim q_\theta(z_t \,|\, o_t)\\
        \text{Dynamics predictor: }\quad &\hat{z}_t  \sim p_\theta(\hat{z}_t \,|\, m_t)\\
        \text{Reward predictor: }\quad &\hat{r}_t    \sim p_\theta(\hat{r}_t \,|\, h_t, z_t)\\
        \text{Continue predictor: }\quad &\hat{c}_t  \sim p_\theta(\hat{c}_t \,|\, h_t, z_t)\\
        \text{Decoder: }\quad &\hat{o}_t             \sim p_\theta(\hat{o}_t \,|\, h_t, z_t) 
\end{align*}
With $o_t$ being the observation at time step $t$, $m_t$ 
the output of the sequence model, and $h_t$ 
the internal state of the S5 layers in the sequence model, which we use as the deterministic part of the world state. $z_t$ is the stochastic part of the latent world state, $a_t$ the action taken, $\hat{z}_t$ the predicted latent state, $\hat{r}_t$ the predicted extrinsic reward, $\hat{c}_t$ the predicted continue signal and $\hat{o}_t$ the decoded observation. \Cref{fig:hieros_structure} shows the overall structure of the world model.

During training, the S5WM processes sequential observations $o_0\cdots o_t$ and actions $a_0\cdots a_{t-1}$. The encoder simultaneously computes the posterior $z_t$ from observation $o_t$. The sequence model predicts the next output $m_t$ and deterministic state $h_t$. The dynamics predictor uses $m_t$ to predict the stochastic prior $\hat{z}_t$. The posterior $z_t$ represents the stochastic state with knowledge of the actual observation $o_t$, while the prior $\hat{z}_t$ is the stochastic state predicted solely from the deterministic sequence model output. During imagination, the actor only has access to the prior $\hat{z}$.

The RSSM encoder in DreamerV3 computes the posterior from both $o_t$ and the deterministic model state $h_t$, which makes the parallel computation of $z_t$ impossible. However, \citet{chen_transdreamer_2022} show, that accurate representation can also be learned by making the posterior $z_t$ independent of $h_t$ and only predict the distribution $p(z_t | o_t)$.


As detailed in \Cref{sec:background_structured_state_space_sequence_models},S5 and S4 layers allow for both parallel sequence modeling and iterative autoregressive next state prediction. This makes them more efficient than RNNs for model training and more efficient than Transformer-based models for imagination. While \citet{deng_facing_2023} propose S4WM, an S4-based world model, our use of S5 layers is advantageous. S5 layers excel in modeling longer-term dependencies and are more memory-efficient. Moreover, the direct accessibility of the latent state $x_t$ in S5 layers enables \hieros to utilize the recurrent model state as the deterministic component of the world state $h_t$. This contrasts with other approaches that rely on using the sequence model output as deterministic world model states \citep{chen_transdreamer_2022, deng_facing_2023, micheli_transformers_2023}, which proves less effective in our experiments (\Cref{sec:appC:state_as_deter}).

We use a modified version of an S5 layer, which was proposed by \citet{lu_structured_2023}. They introduce a reset mechanism, which allows setting the internal state $h_t$ to its initial value $h_0$ during the parallel sequence prediction by also passing the continue signal $c_0\cdots c_n$ to the sequence model. This allows the actor to train on trajectories that span multiple episodes without leaking information from the end of one episode to the beginning of the next. We employ the same mechanism for our S5WM, as trajectories spanning episode borders are commonly encountered during training. It is unclear if the S4WM proposed in \citet{deng_facing_2023} faced the same challenge and if and how they solved it.

The sequence model of our S5WM uses a stack of multiple S5 blocks, as depicted in \Cref{fig:hieros_structure}. The design of this block is inspired by the deep sequence model architecture proposed by \citet{smith_simplified_2023} but we selected a different norm layer, dropout and activation function. The number of blocks used is listed in \Cref{sec:appB}.
All parts of S5WM are optimized jointly using a loss function that follows the loss function of DreamerV3:
\begin{align}
    &\mathcal{L}(\theta) = \mathcal{L}_{pred}(\theta) + \alpha_{dyn}\cdot\mathcal{L}_{dyn}(\theta) + \alpha_{rep}\cdot\mathcal{L}_{rep}(\theta) \\
    &\mathcal{L}_{pred}(\theta) = -\ln(\,p_\theta(r_t|h_t, z_t)) -\ln(p_\theta(c_t|h_t, z_t)) \\ & \quad\quad\quad\,\,\,\,\quad \, -\ln(p_\theta(o_t|h_t, z_t)) \nonumber \\
    &\mathcal{L}_{dyn}(\theta) = \max(1, \text{KL}\left[\,\text{sg}(\,q_\theta(z_t|o_t)) \,||\, p_\theta(z_t| m_t)\right]) \\
    &\mathcal{L}_{rep}(\theta) = \max(1, \text{KL}\left[\,q_\theta(z_t|o_t) \,||\, \text{sg}(\,p_\theta(z_t| m_t))\right]) 
\end{align}
For $\mathcal{L}_{dyn}$ and $\mathcal{L}_{rep}$ we use the method of free bits introduced by \citet{kingma2016improved} and used in DreamerV3 \citep{hafner_mastering_2023} to prevent the dynamics and representations from collapsing into easily predictable distributions. $\alpha_{dyn}$ and $\alpha_{rep}$ are hyperparameters that control the influence of the dynamics and representation loss.

\subsection{Efficient Time-balanced Sampling}
\label{sec:efficient_time_balanced_sampling}

When interacting with the environment, \hieros collects observations, actions, and rewards in an experience dataset. After a fixed number of interactions, trajectories are sampled from the dataset in order to train the world model, actor, and subgoal autoencoder. DreamerV3 \citep{hafner_mastering_2023} uses a uniform sampling. This, however, leads to an oversampling of the older entries of the dataset, as the iterative uniform sampling can select these entries more often than newer ones.
As explained in \Cref{sec:introduction} \citet{robine_transformer-based_2023} solve this using a time-balanced replay buffer with a $O(n)$ sampling runtime complexity with $n$ being the size of the replay buffer.

We propose an efficient time-balanced sampling method (ETBS), which produces similar results with $O(1)$ time complexity. 
When iteratively adding elements to the experience replay dataset and afterward sampling uniformly, the expected number of times $N_{x_i}$ an element $x_i$ has been drawn after $n$ iterations is 
\begin{align}
    \mathbb{E}({N_{x_i}}) = \frac{1}{i} + \frac{1}{i + 1} + \cdots + \frac{1}{n} = H_n - H_i
\end{align}
with $H_i$ being the $i$-th harmonic number. 
The probability of sampling $x_i$, $i=1,...,n$, is
\begin{align}
p_i = \frac{1}{n} \cdot \mathbb{E}({N_{x_i}}) = \frac{H_n - H_i}{n} \approx \frac{\ln(n) - \ln(i)}{n}
\end{align}
We use the approximation $H_x \approx \ln(x)$ to remove the need to compute harmonic values. The idea is, to compute the CDF of this probability distribution between 0 and $n$ to transform samples from the imbalanced distribution into uniform samples via probability integral transformation \cite{david1948probability}. How we derive the CDF of this distribution is shown in \Cref{sec:appD:sampling}. With the CDF we compute the ETBS probabilities as follows:
\begin{align}
    p_{etbs}(x) = CDF(\,p(x)\,) \cdot \tau + p(x) \cdot (1 - \tau)
\end{align}
with $p$ being the original sampling distribution, $CDF$ being the CDF of the original distribution and $\tau$ being a temperature hyperparameter that controls the influence of the original distribution on the overall distribution. We set $\tau$ to 0.3 in our experiments. Empirically, a slight oversampling of earlier experiences seems to have a positive influence on the actor performance. The time complexity of this sampling method is $O(1)$, as the CDF can be precomputed and the sampling is done in constant time.
%
For further details, see \Cref{sec:appD:sampling}.
\section{Experiments}
\label{sec:experiments}

We evaluate the performance of \hieros on the Atari100k test suite \citep{bellemare2013arcade}, which contains a wide range of games with different dynamics and objectives. In each of these environments, the agent is only allowed 100k interaction with the environment, which amounts to roughly 2 hours of total gameplay. We evaluate \hieros on a subset of 25 of those games. We compare \hieros to the following baselines: DreamerV3 \citep{hafner_mastering_2023}, IRIS \citep{micheli_transformers_2023}, TWM \citep{robine_transformer-based_2023}, and SimpPLe \citep{kaiser2019model}. 
SimPLe trains a policy on the direct pixel input of the environment using PPO \citep{schulman2017proximal}, while TWM, IRIS and DreamerV3 train a policy on imagined trajectories of a world model. TWM and IRIS use a Transformer-based world model. IRIS, however, trains the agent not in the latent space of the world model but in the decoded state space of the environment, which is one of the main differences to other approaches that leverage a Transformer-based world model (e.g. TWM). \citet{ye2021mastering} propose EfficientZero, which holds the current absolute state of the art in the Atari100k test suite. However, this method relies on look-ahead search during policy inference and is thus not comparable to the other methods.

\begin{figure*}[t]   
    \centering
    \includegraphics[width=0.7\textwidth]{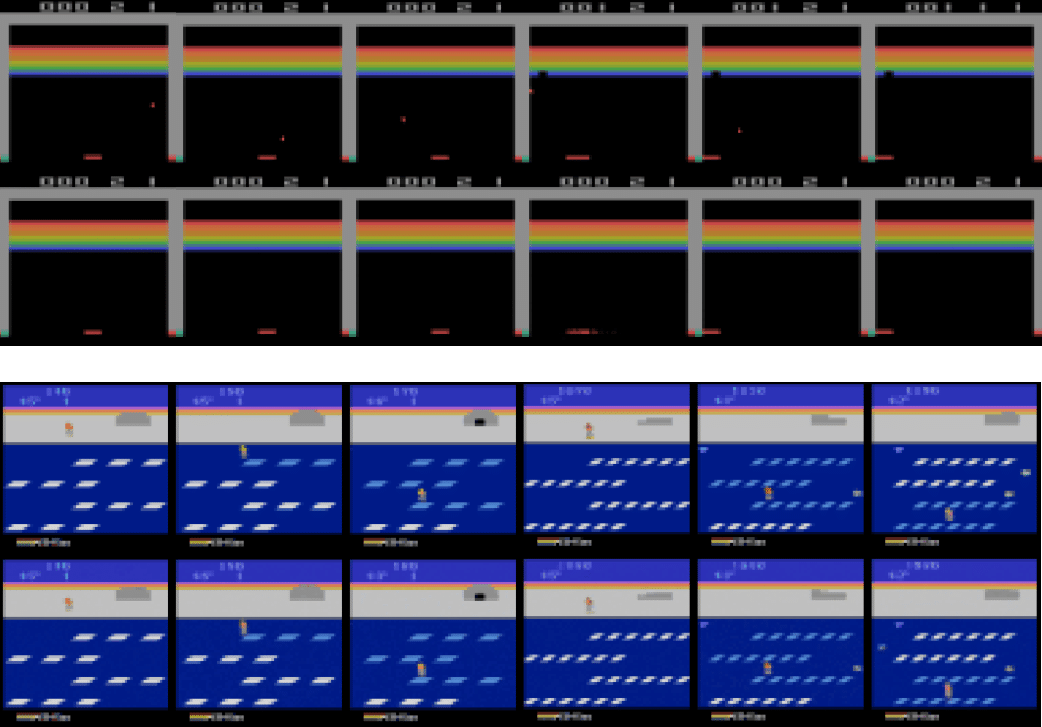}
    \caption{Trajectories for Breakout (top) and Frostbite (bottom). For each, the upper frame is the image observed in the environment and the lower frames are the imagined trajectories of the S5WM of the lowest level subactor.
    \vspace{-0.3cm}}
\label{fig:trajectories}
\end{figure*}

Comparing our S5WM against the S4WM proposed by \citet{deng_facing_2023} would be interesting, as both models are based on structured state spaces. However, \citet{deng_facing_2023} only report results on their proposed set of memory testing environments and their code base is not public yet. So we are not able to compare our results directly to theirs. We do however compare our S5WM to the RSSM used in DreamerV3 in \Cref{sec:imagined_trajectories}.
The used computation resources, implementation details and a link to the source code can be found in \cref{sec:resources}. We use the hyperparameters as specified in \Cref{sec:appB} for all experiments.

\subsection{Results for the Atari100k Test Suite}

\begin{table}[t]
  \centering
  \resizebox{1.0\columnwidth}{!}{
  \begin{tabular}{|l|cccc|}
  \hline &&&& \\[-2ex]
      Task & Mean & Median & IQM & Optimality Gap \\\hline &&&& \\[-2ex]
      Random            & 0            & 0            & 0                   & 100 \\ 
      Human          & 100            & 100            & 100                   & 0\\\hline &&&& \\[-2ex]
      SimPLe          & 34            & 11            & 13                   & 73 \\ 
      TWM          & 96            & 50            & 46                   & 52   \\ 
      IRIS          & 105            & 29            & 50                   & 51   \\ 
      DreamerV3          & 112            & 49            & N/A                   & N/A     \\ 
      Hieros (ours)          & \textbf{120}            & \textbf{56}            & \textbf{53}                   & \textbf{49}   \\ \hline

  \end{tabular}
  }
  \caption{Aggregate scores of \hieros and baselines on the Atari100k test suite. Higher scores for Mean, Median and IQM are better. For Optimality Gap, lower scores are better. DreamerV3 does not report the scores for IQM or Optimality Gap. We show the best results for each row in \textbf{bold} font.
  \vspace{-0.2cm}}
  \label{tab:small_atari100k_scores}
\end{table}

All scores are the average of three runs with different seeds. To aggregate the results, we compute the normalized human score \citep{bellemare2013arcade}, which is defined as $(score_{agent} - score_{random}) / (score_{human} - score_{random})$. We show the achieved aggregated mean and median normalized human score in \Cref{tab:small_atari100k_scores}. The full table with scores for all games can be found in \Cref{sec:full_table}.

Our model achieves a new state of the art in regard to the mean and median normalized human score. We also achieve a new state of the art in regard to the achieved reward on 9 of the 25 games. Adhering to \citet{agarwal2021deep}, we also report the optimality gap and the interquartile mean (IQM) of the human normalized scores and achieve state-of-the-art results in both of those metrics. \hieros outperforms the other approaches while having significant advantages with regard to runtime efficiency during training and inference, as well as resource demand. For training on a single Atari game for 100k steps, TWM takes roughly 0.8 days on a A100 GPU, while DreamerV3 takes 0.5 days and IRIS takes 7 days. Hieros takes roughly 14 hours $\approx$ 0.6 days and is thus significantly faster than IRIS while being on par with DreamerV3 and TWM.

Significant improvements were achieved in Frostbite, JamesBond, and PrivateEye. These games feature multiple levels with changing dynamics and reward distributions. In order for the S5WM to learn to simulate these levels, the actor needs to employ a sufficient exploration strategy to discover these levels. This shift in the state distribution poses a challenge for many imagination-based approaches \citep{micheli_transformers_2023}. \hieros is able to overcome this challenge by using the proposed subgoals on different time scales to guide the agent towards the next level. We show in \Cref{sec:app_visual} different proposed subgoals which guide the lower level actor to finding the way to the next level by building the igloo in the upper right part of the image. 

\hieros seems to perform significantly worse than other approaches in Breakout or Pong, which feature no significant shift in states or dynamics. It seems as if the hierarchical structure makes it harder to grasp the relatively simple dynamics of these games, as the dynamics remain the same across all time abstractions. This is backed by our findings in \Cref{sec:appC:model_hierarchy_depth} that \hieros with only one subactor performs significantly better on Breakout. We also show empirically in \Cref{sec:imagined_trajectories} that using the S5WM also seems to deteriorate the performance of \hieros in those games compared to the RSSM used for DreamerV3. \Cref{fig:subgoals} shows some proposed subgoals for Breakout. The subgoals seem only to propose to increase the level score, which is does not provide the lower level agent any indication on how to do so. So the lower level actor is not able to benefit from the hierarchical structure in these games. 

\begin{figure}[t]  
    \centering
    \includegraphics[width=0.45\textwidth]{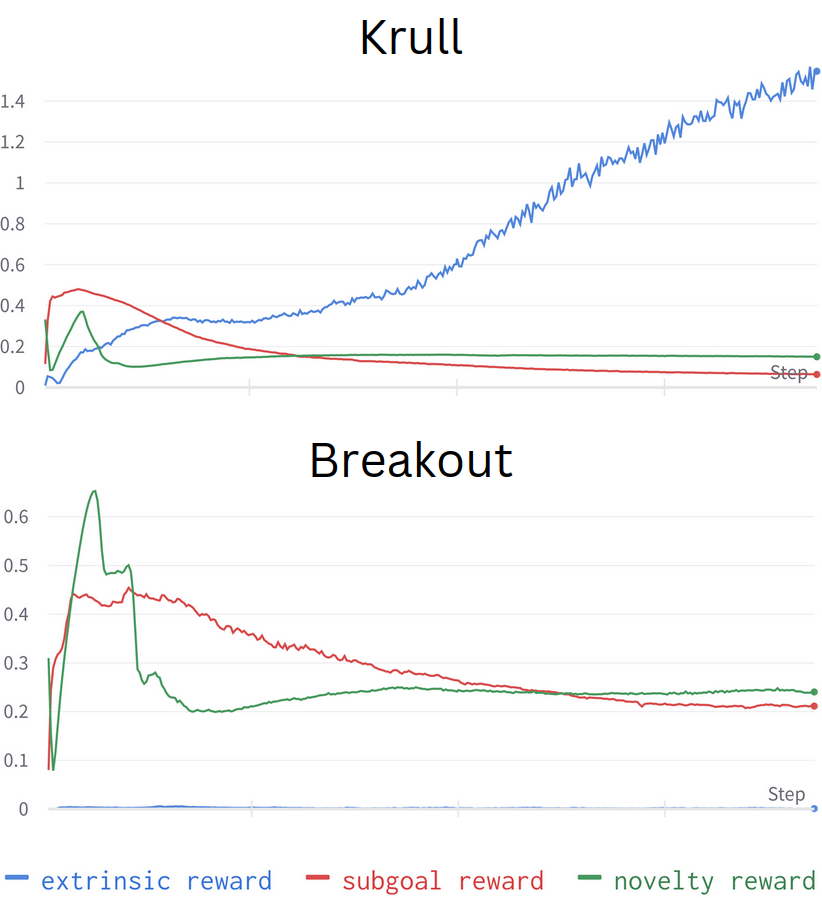}
    \vspace{-0.1cm}
    \caption{Extrinsic, subgoal, and novelty rewards per step for Krull (top) and Breakout (bottom) for the lowest level subactor.
    \vspace{-0.05cm}}
    \label{fig:rewards}
\end{figure}

\Cref{fig:rewards} shows the partial rewards $r_{extr}, r_{nov}$, and $r_{sg}$ for the lowest level subactor for Breakout and Krull, another game featuring multiple levels similar to Frostbite. As can be seen, the extrinsic rewards for Breakout are very sparse and do not provide any indication on how to increase the level score. So the subactor learns to follow the subgoals from the higher levels more closely instead, which in the case of Breakout or Pong does not lead to a better performance. In Krull however, the extrinsic rewards are more frequent and provide a better indication on how to increase the level score. So the subactor is able to learn to follow the subgoals from the higher levels more loosely, treating them more like a hint, and is able to achieve a better performance.

\subsection{Imagined Trajectories}
\label{sec:imagined_trajectories}

In \Cref{fig:trajectories}, we show an observed trajectory for the games Frostbite and Breakout alongside the imagined trajectories of the S5WM of the lowest level subactor. As can be seen, the world model is not able to predict the movement of the ball in breakout. This indicates that the model is not able to model the very small impact of the ball movement to the change in the image. We found that during random exploration at the beginning of the training the events where the ball bounces back from the moving plateau are very rare and most of the time the ball is lost before it can bounce back. This makes it difficult for the world model to learn the dynamics of the ball movement and their impact on the reward distribution. We make similar observations for the game Pong. 

S4-based models are shown to perform worse than Transformer models on short term sequence modeling tasks \citep{zuo2022efficient, mehta2022long, dao2022hungry} while excelling on long term modeling tasks. This could also give some reasoning why our S5WM seems to perform worse on Breakout or Pong compared to Frostbite or Krull.
Freeway poses a similar challenge, with sparse rewards that are only achieved after a complex series of environment interaction. However, unlike with Breakout and Pong, in Freeway \hieros is able to profit from its hierarchical structure in order to guide exploration. An example of this can be seen in \Cref{fig:subgoals}, where the subgoals are able to guide the agent across the road.
S5WM is able to accurately capture the multiple levels of Frostbite, despite only having access to train on these levels after discovering them, which usually happens after roughly 50k interactions. As the reaching of the next level is directly connected to a large increase in rewards, the S5WM is able to correctly predict the next level after only a few interactions. This is also reflected in the significantly higher reward achieved by \hieros.

\begin{figure}[t]  
    \centering
    \includegraphics[width=0.45\textwidth]{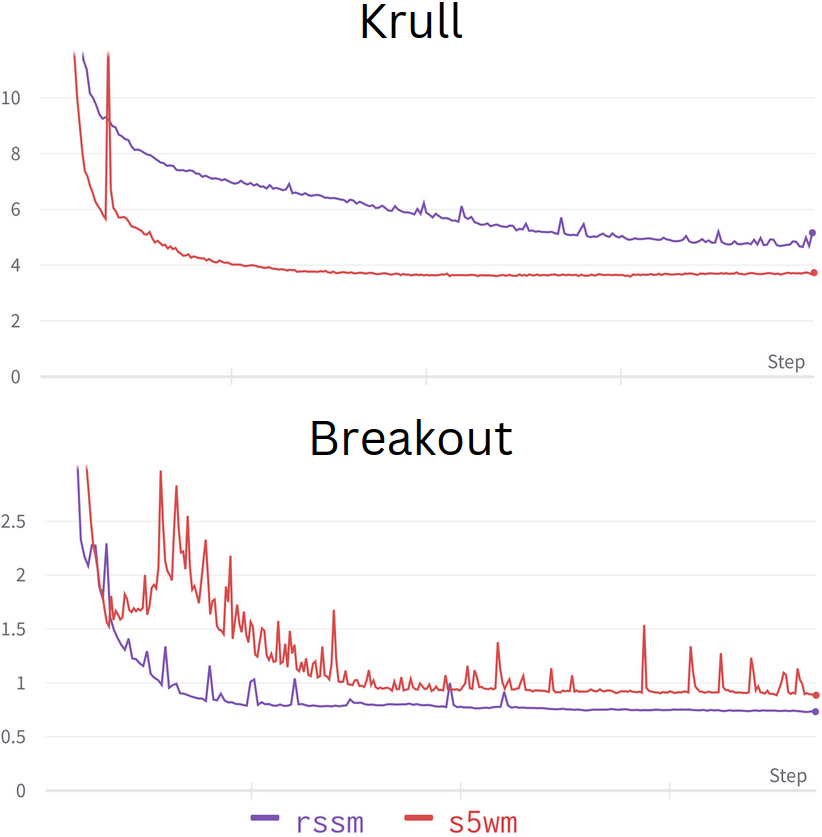}
    \vspace{-0.1cm}
    \caption{World model losses for the S5WM and RSSM for Krull and Breakout. The S5WM is able to achieve an overall lower world model loss compared to the RSSM for Krull, while those roles are reversed for Breakout.
    \vspace{-0.3cm}}
\label{fig:losses}
\end{figure}

To directly compare the influence of our S5WM architecture to the RSSM used in DreamerV3, we replace the S5WM with an RSSM and train \hieros on four different games. The results are shown in \Cref{ablation:rssm}. The RSSM is not able to achieve the same performance as the S5WM for Krull, Battle Zone and Freeway, but is slightly better compared to using the S5WM for Breakout. In \Cref{fig:losses} we show the world model losses for S5WM and RSSM for Krull and Breakout. The S5WM consistently achieves lower overall loss than the RSSM in Krull, whereas this trend reverses for Breakout. Given the substantially higher absolute loss values for the more complex Krull compared to Breakout, we infer that the larger S5WM excels in environments with complex dynamics and substantial input distribution shifts. In contrast, the smaller RSSM is better suited for environments with simple dynamics and a stable input distribution.

As both models are trained with the same number of gradient updates, it makes sense that the larger model has difficulties matching the performance of the smaller model in non-shifting environments with simple dynamics. So a smaller S5WM might achieve better results for Breakout. We test this hypothesis in \Cref{app:smol_s5}. \citet{lu_structured_2023} and \citet{deng_facing_2023} find that structured state space models generally surpass RNNs in terms of memory recollection and resilience to distribution shifts, which we can confirm with our results.
Moreover, we perform further ablations in \Cref{sec:appC}.


\section{Conclusion}
\label{sec:conclusion}

In this paper, we introduce the \hieros architecture, a multilayered goal conditioned hierarchical reinforcement learning (HRL) algorithm with hierarchical world models, an S5 layer-based world model (S5WM) and an efficient time-balanced sampling (ETBS) method which allows for a true uniform sampling over the experience dataset. We evaluate \hieros on the Atari100k test suite \citep{bellemare2013arcade} and achieve a new state of the art mean human normalized score for model-based RL agents without look-ahead search. The option to decode proposed subgoals gives some explainability to the actions taken by \hieros. A deeper evaluation on which subgoals are proposed and how the lower level workers are able to achieve these subgoals is left for future research.
The S5WM poses multiple improvements compared to RNN-based world models \citep{hafner_mastering_2023} (i.e. efficiency during training, prediction accuracy during imagination) and Transformer-based world models \citep{chen_transdreamer_2022,robine_transformer-based_2023,micheli_transformers_2023} (i.e. prediction accuracy and efficiency during imagination). A thorough comparison between our S5WM and the S4WM proposed by \citet{deng_facing_2023} remains for future research. We delve deeper into further directions for future research in \Cref{sec:future_work}.

We would also like to point out that the model-free SR-SPR recently proposed by \citet{d2022sample} achieves superior scores on the Atari100k benchmark by increasing the replay train ratio and regular resets of all model parameters. They tackle the often observed inability of RL agents to learn new behavior after having been already trained for some time in an environment. However, since this approach is not model based and does not train in imagination, we did not include it into our comparison. Nonetheless, scaling the train ratio and employing model resets for the actor/critic or the S5WM of \hieros are promising directions for future research.

%
With our work, we hope to provide 
a new perspective on the field of 
HRL 
and to inspire future research in this field.

\newpage

\section*{Reproducibility Statement}
We describe all important architectural and training details in \Cref{sec:methodology} and provide the used hyperparameters in \Cref{sec:appB}. We provide the source code in the supplementary material and under the following link: \repourl. The material also gives an explanation on how to install and use the Atari100k benchmark for reproducing our results. The computational resources we used for our experiments are described in \Cref{sec:resources}.

\section*{Impact Statement}

Autonomous agents pose many ethical concerns, as they are able to act in the real world and can cause harm to humans. In our work, we only use simulated environments and do not see any potential for misuse of our work. We propose a new world model architecture which could be used to train agents in imagination rather than in the real world. This could be used to train agents for real-world applications, such as autonomous driving, without the need to train them in the real world. This could reduce the risk of harm to humans and the environment.

\bibliography{hieros}
\bibliographystyle{icml2024}

\appendix

\onecolumn
\section{Background}
\label{sec:background}

\subsection{Learning a World Model in DreamerV3}
\label{sec:background_world_models}
Our method is build on top of DreamerV3 \citep{hafner_mastering_2023}. DreamerV3 learns a world model in a latent space in order to increase efficiency:
\begin{align}
    (h_t, z_t) = \text{RSSM}(h_{t-1}, a_{t-1}, o_t) 
\end{align}
with $h_t$ being the deterministic part of the latent state and $z_t$ being the stochastic part of the latent state. The world model is trained to predict the next latent state $(h_{t+1}, z_{t+1})$ given the world model state $(h_t, z_t)$ and the current action $a_t$. Specifically, the world model learns a distribution $p_\theta(z_t | h_t, o_t)$ predicting the stochastic state from the last deterministic state and an observation $o_t$ and another distribution $q_\theta(z_t | h_t)$ predicting the stochastic state purely from the last deterministic state. $q_\theta$ is used during imagination. It then predicts the reward $r$, the continue signal $c$ and the decoded observation $o$ given the current world state, which enables DreamerV3 to train an actor critic entirely on imagined trajectories in latent space. The world model is trained with a loss function that is a weighted sum of the loss of the dynamic, observation, continue, and reward prediction.

The RSSM dynamic prediction model uses a GRU \citep{cho2014learning} to predict the next deterministic state $h_{t+1}$ and a discrete categorical distribution to sample $z_{t+1}$. There are several works that replace the GRU with different, more complex models \citep{chen_transdreamer_2022,robine_transformer-based_2023,deng_facing_2023}.

For imagining the trajectories, the world model starts with an initial observation $o_0$ and an initial state $h_0$. It then computes the first world state $(h_0, z_0)$. The agent interacts with this model, which simulates the original environment:
\begin{align}
    a_t &= \pi(h_t, z_t) \\
    (h_{t+1}, z_{t+1}) &= \text{RSSM}(h_t, a_t, z_{t+1})
\end{align}

\subsection{Hierarchical Reinforcement Learning}
\label{sec:background_hierarchical_reinforcement_learning}

Hierarchical models have been shown to be a powerful tool in RL 
\citep{dayan1992feudal,parr1997reinforcement, sutton1999between}. The idea is to break down a complex task into easily achievable subtasks. This subtask definition can be done manually \citep{tessler2017deep} or automatically \citep{li_hierarchical_2022,nair_hierarchical_2019,kujanpaa_hierarchical_2023}. This typically involves learning a high level actor that works at larger timescale and a low level actor that executes proposed subgoals \citep{hafner_deep_2022,nachum_data-efficient_2018,jiang_language_2019,nachum_multi-agent_2019}:
\begin{align}
    g_t &= \pi_{high}(s_t) \\
    a_t &= \pi_{low}(s_t, g_t)
\end{align}
with $s_t$ being the current environment state, $g_t$ being the proposed subgoal and $a_t$ being the action taken by the low level actor. The low level actor is often encouraged to fulfill the subgoal by adding a subgoal reward $r_{g}$ to the extrinsic reward $r_{extr}$. This subgoal reward is often a function of the distance between the current state and the proposed subgoal \citep{hafner_deep_2022,nachum_multi-agent_2019}.

\subsection{Structured State Space Sequence Models}
\label{sec:background_structured_state_space_sequence_models}

Structured state space sequence models (S4) were initially introduced by \citet{gu_efficiently_2022} as a sequence modeling method that is able to achieve superior long-term memory tasks than Transformer-based models while having a lower runtime complexity ($O(n^2)$ for the attention mechanism of Transformers and $O(n \log n)$ for S4, $n$ being the sequence length). State Space models are composed of four matrices: $A$, $B$, $C$ and $D$. They take in a signal $u(t)$ and output a signal $y(t)$:
\begin{align}
    x(t+1) &= Ax(t) + Bu(t) \\
    y(t) &= Cx(t) + Du(t)
\end{align}
with $x(t)$ being the state of the model at time $t$. The matrices $A$, $B$, $C$, and $D$ are learned during training.
\citet{gu_efficiently_2022} propose various techniques to increase stability, performance and training speed of these models in order to model long sequences. They utilize special HIPPO initialization matrices for this. Another major advantage of S4 layers over Transformer models is, besides their better runtime efficiency, that they can be used both as a recurrent model, which allows for fast autoregressive single step prediction, and as a convolutional model, which allows for fast parallel sequence modelling. \citet{smith_simplified_2023} propose a simplified version of S4 layers (S5) that is able to achieve similar performance while being more stable and easier to train. Their version utilizes parallel scans and different matrix initialization in order to further boost the parallel sequence prediction and runtime performance.
\citet{lu_structured_2023} propose a resettable version of S5 layers, which allow resetting the internal state $x(t)$ during the parallel scans, in order to apply S5 layers in a RL setting where the state input sequence might span episode borders.

\section{Further Related Work}
\label{sec:related_work}

\subsection{Hierarchical Reinforcement Learning}
\label{sec:related_work_hierarchical_reinforcement_learning}

Hierarchical reinforcement learning (HRL) is a field of RL 
that breaks down complex tasks in time and state abstracted subproblems on multiple time scales \citep{hutsebaut-buysse_hierarchical_2022, sutton1999between}. This allows the agent to learn subtasks on different time scales and to reuse these subtasks in different contexts. This is especially useful in sparse reward environments, where the agent can learn subtasks that are easier to solve and then combine them to solve the overall task \citep{nair_hierarchical_2019}. \citet{nachum_why_2019} show that one of the main benefits of HRL is improved exploration, more than inherent hierarchical structures of the problem task itself or larger model sizes. \citet{lecun2022path} argue that HRL is a promising direction for future research in 
RL, 
as complex dependencies often are only discoverable by abstraction, as learned skills are often reusable in different contexts. Also, they show that HRL has deep rooting in human cognition.

Goal-conditioned 
HRL 
\citep{florensa2017stochastic, nachum_data-efficient_2018, nachum_near-optimal_2019, nachum_multi-agent_2019} is a subfield of HRL that uses goal-conditioned policies to learn subtasks. The agent learns a policy that takes a goal as input and outputs actions that lead to the goal. The agent can then learn a policy that takes a goal as input and outputs a subgoal that leads to the goal. This allows the agent to learn subtasks on different time scales and to reuse these subtasks in different contexts. The higher order policy proposes subgoals in frequent intervals that the lower order policy has to fulfill, which is often incentivized by giving an intrinsic reward to the lower level policy \citep{hafner_deep_2022,rosete-beas_latent_nodate}.
\citet{hafner_deep_2022} combine a hierarchical policy with a world model, building on the DreamerV2 architecture \citep{hafner_mastering_2022}. They show that the combination of a hierarchical policy and a world model outperforms the original DreamerV2 model on several tasks. The low level policy receives only subgoal rewards in this architecture, while the higher level policy receives the actual task reward. This is similar to our approach, but we use a different architecture for the world model, and we let the higher level policies learn a separate world model.

\subsection{World Models}
\label{sec:related_work_world_models}

Environment interactions are typically expensive to train an RL 
agent. E.g., in robotic applications it is impossible to let the agent interact with the real environment, as this would be too expensive and potentially dangerous. Therefore, it is desirable to train the agent in a simulated environment \citep{ha2018world,poudel_contrastive_2022,hafner_dream_2020,hafner_mastering_2022,hafner_mastering_2023,hafner_learning_nodate}. \citet{ha2018world} introduced learning an RNN-based model of the environment and using this model to train the agent. This allows the agent to learn from simulated data, which is much cheaper than learning from real environment interactions and can in principle be generated in an infinite amount. \citet{hafner_dream_2020} introduced Dreamer, a model-based RL 
agent that learns a world model based on PlaNet \citep{hafner_learning_nodate} and uses this world model to train the agent. PlaNet uses an RNN architecture which predicts the next world state from the last world state and the next action from the learned policy. It uses an RNN-based architecture. To increase computational efficiency, all learning is done in a compact latent state. They show that Dreamer outperforms state of the art model-free 
RL 
agents on several tasks. 

\citet{hafner_mastering_2022} introduced DreamerV2, an improved version of Dreamer featuring a discrete stochastic latent state. They show that DreamerV2 outperforms Dreamer on several tasks. With DreamerV3 \citep{hafner_mastering_2023} they propose some additional improvements with which they were able to solve the Minecraft Diamond challenge \citep{kanitscheider2021multi} without any pretraining. DreamerV3 still uses a PlaNet-based world model, but the model states are composites of a discrete valued stochastic and a continuous valued deterministic part.

Several authors propose improvements to this architecture, mostly by proposing improvements to the used world model. \citet{chen_transdreamer_2022} propose replacing the RNN with a Transformer model that takes in a context $(s_0, a_0),...,(s_n, a_n)$ of state action pairs $(s_i, a_i)$ in order to compute the next world state. \citet{robine_transformer-based_2023} propose a similar architecture, but in contrast to the previous work, the Transformer model is not used during inference, which makes their model more computationally efficient. \citet{deng_facing_2023} propose S4WM, utilizing S4 layers for the next state prediction. Since S4 layers can be used both for predicting sequences in parallel and predicting only the next value in an RNN like fashion, their model also proved to be more computationally efficient than the Transformer-based architectures and outperforms them in memorization capabilities.



\section{Full Atari100k Benchmark}
\label{sec:full_table}

\begin{table*}[h]
  \centering
  \resizebox{0.9\textwidth}{!}{
  \begin{tabular}{|l|rr|rrrrr|}
  \hline
      Task & Random & Human & SimPLe & TWM & IRIS & DreamerV3 & Hieros (ours) \\\hline
      Alien           & 228          & 7\,128           & 617                  & 675        & 420       & \textbf{959}          & 828 \\
      Amidar          & 6            & 1\,720           & 74                   & 123         & 143          & 139        & 127 \\
      Assault         & 222          & 742            & 527                  & 683          & 1\,524        & 706       & \textbf{1\,764} \\
      Asterix         & 210          & 8\,503           & \textbf{1\,128}        & 1\,117        & 854       & 932         & 899 \\
      BankHeist       & 14           & 753            & 34                   & 467        & 53        & \textbf{649}         & 177 \\
      Battle Zone     & 2\,360         & 37\,188          & 4\,031                 & 5\,068        & 13\,074         & 12\,250          & \textbf{15\,140} \\
      Boxing          & 0            & 12             & 8                    & 78         & 70         & \textbf{78}        & 65 \\
      Breakout        & 2            & 30             & 16                   & 20         & \textbf{84}          & 31       & 10 \\
      Chop.Command    & 811          & 7\,388           & 979                  & \textbf{1\,697}          & 1\,565       & 420          & 1\,475 \\
      CrazyClimber    & 10\,780        & 35\,829          & 62\,584                & 71\,820          & 59\,324          & \textbf{97\,190}           & 50\,857 \\
      DemonAttack     & 152          & 1\,971           & 208                  & 350        & \textbf{2\,034}       & 303         & 1\,480 \\
      Freeway         & 0            & 30             & 17                   & 24       & \textbf{31}       & 0       & \textbf{31} \\
      Frostbite       & 65           & 4\,335           & 237                  & 1\,476       & 259       & 909        & \textbf{2\,901} \\
      Gopher          & 258          & 2\,412           & 597                  & 1\,675          & 2\,236       & \textbf{3\,730}        & 1\,473 \\
      Hero            & 1\,027         & 30\,826          & 2\,657                 & 7\,254         & 7\,037       & \textbf{11\,161}        & 7\,890 \\
      JamesBond       & 29           & 303            & 100                  & 362      & 463         & 445        & \textbf{939} \\
      Kangaroo        & 52           & 3\,035           & 51                   & 1\,240         & 838        & 4\,098          & \textbf{6\,590} \\
      Krull           & 1\,598         & 2\,666           & 2\,205                 & 6\,349         & 6\,616        & 7\,782       & \textbf{8\,130} \\
      KungFuMaster    & 258          & 22\,736          & 14\,862                & \textbf{24\,555}          & 21\,760           & 21\,420          & 18\,793 \\
      Ms.Packman      & 307          & 6\,952           & 1\,480                 & 1\,588         & 999          & 1\,327       & \textbf{1\,771} \\
      Pong            & -21          & 15             & 13                   & \textbf{18}         & 15         & \textbf{18}        & 5 \\
      PrivateEye      & 25           & 69\,571          & 35                   & 86         & 100        & 882         & \textbf{1\,507} \\
      Qbert           & 164          & 13\,455          & 1\,289                 & 3\,331        & 746       & \textbf{3\,405}       & 770 \\
      RoadRunner      & 12           & 7\,845           & 5\,641                 & 9\,109          & 9\,615       & 15\,565       & \textbf{16\,950} \\
      Seaquest        & 68           & 42\,055          & 683         & \textbf{774}        & 661        & 618       & 560 \\\hline &&&& \\[-2ex]
      Mean            & 0            & 100            & 34                   & 96         & 105        & 112        & \textbf{120} \\ 
      Median          & 0            & 100            & 11                   & 50         & 29         & 49         & \textbf{56} \\
      IQM            & 0            & 100            & 13                   & 46         & 50        & N/A        & \textbf{53} \\ 
      Optimality Gap          & 100            & 0            & 73                   & 52         & 51         & N/A         & \textbf{49} \\ \hline
  \end{tabular}
  }
  \caption{Scores of \hieros and baselines on the Atari100k test suite. Higher scores for Mean, Median and IQM are better. For Optimality Gap, lower scores are better. DreamerV3 does not report the scores for IQM or Optimality Gap. We show the best results for each row in \textbf{bold} font.
  \vspace{-0.2cm}}
  \label{tab:atari100k_scores}
\end{table*}

\newpage
\section{Visualization of Proposed Subgoals}
\label{sec:app_visual}

\begin{figure*}[ht]  
    \centering
    \includegraphics[width=0.7\textwidth]{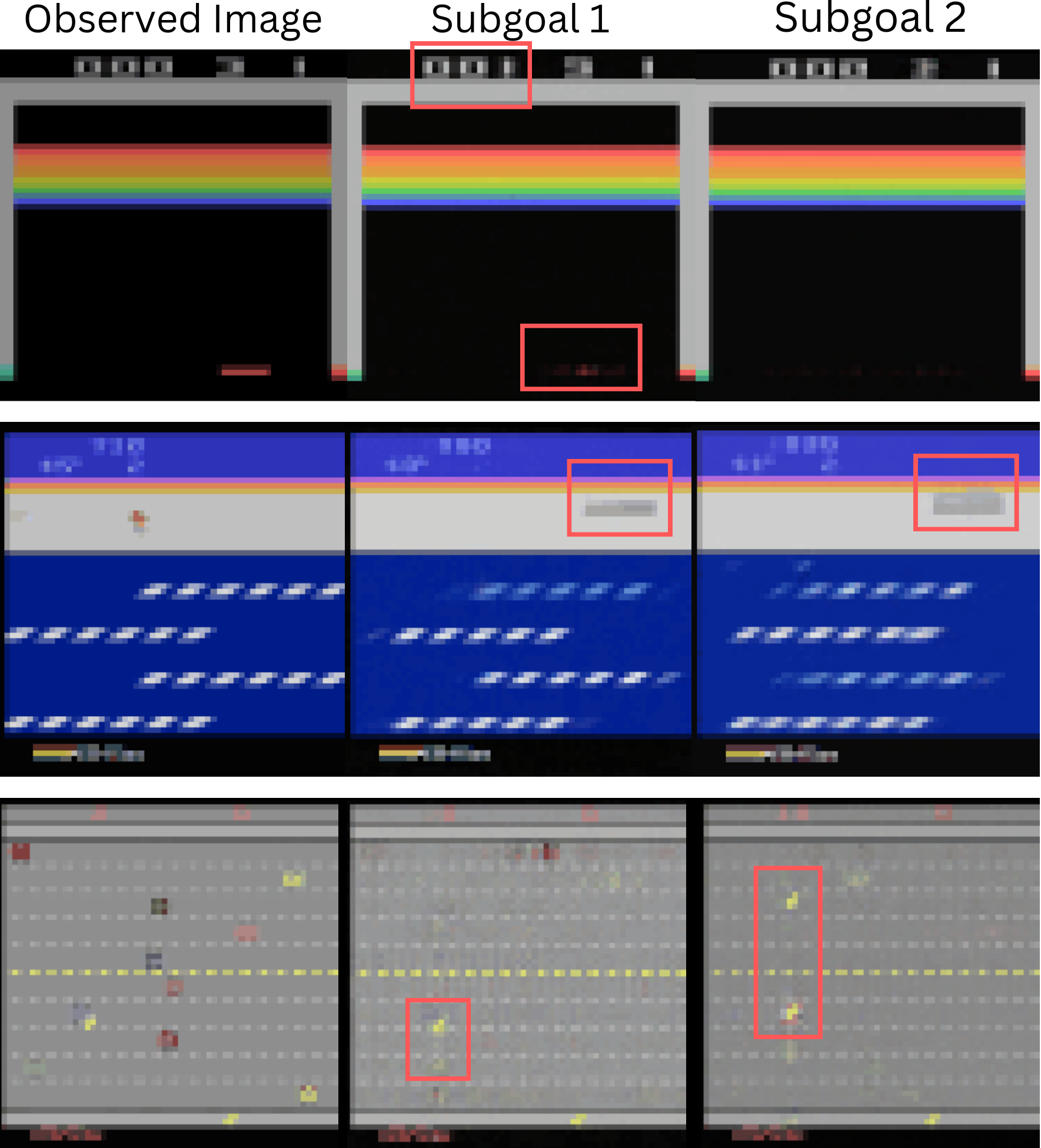}
    \caption{Proposed subgoals for Breakout (top row), Frostbite (middle row), and Freeway (bottom row). The left most frame is the original observation from the environment, and the following frames are the proposed subgoals from the higher level actor. For Breakout, the subgoals are only to increase the level score (marked with the red rectangles) and the ball is not simulated at all, while for Frostbite the subgoals guide the actor towards building up the igloo in the upper right part of the image in order to advance to the next level (red rectangles). For Freeway, which also features a single level and sparse rewards, the subgoals are much more meaningful than for Breakout and guide the actor to move across the road (red rectangles).
    \vspace{-0.2cm}}
    \label{fig:subgoals}
\end{figure*}

\Cref{fig:subgoals} shows the proposed subgoals for one observation in Frostbite and one observation in Breakout.

\newpage\newpage
\begin{samepage}
    
    \section{Hyperparameters}
    \label{sec:appB}
    \begin{table*}[ht!]
        \centering
        \begin{tabular}{lc}
            \hline
            Training parameters:                        &           \\\hline
            learning rate                               & $10^{-4}$ \\
            weight decay                                & 0         \\
            optimizer                                   & AdamW     \\
            learning rate scheduler                     & None      \\
            warmup episodes                             & 1000      \\
            ETBS temperature $\tau$                     & 0.3       \\
            batch size                     & 16       \\
            trajectory length for training                     & 64       \\
            imagination horizon                     & 16       \\
            &           \\\hline
            Hierarchy parameters:                       &           \\\hline
            hierarchy layers                            & 3         \\
            subgoal proposal intervals $k$              & 4         \\
            extrinsic reward weight                     & 1         \\
            subgoal reward weight                       & 0.3       \\
            novelty reward weight                       & 0.1       \\
            subgoal shape                               & 8x8       \\
            &           \\\hline
            World Model parameters:                     &           \\\hline
            S5 model dimension                          & 256       \\
            S5 state dimension                          & 128       \\
    Number of HIPPO-N initialization blocks $J$ & 4         \\
    Number of S5 blocks & 8 \\
    dynamic loss weight $a_{dyn}$               & 0.5       \\
    representation loss weight $a_{rep}$        & 0.1       \\
    $h_t$ dimension                             & 256       \\
    $z_t$ dimension                             & 32x32     \\
    MLP units                                   & 256       \\
    $g_\psi$ KL loss weight $\beta$                                   & 0.5       \\
    &           \\\hline
    Total parameters                            & 37.1 M   \\\hline
\end{tabular}
\end{table*}
\end{samepage}

\section{Computational Resources and Implementation Details}
\label{sec:resources}

In our experiments we use a machine with an NVIDIA A100 graphics card with 40 GB of VRAM, 8 CPU cores and 32 GB RAM. Training \hieros on one Atari game for 100k steps took roughly 14 hours in our setup.

We base our implementation on the Pytorch implementation of DreamerV3 \citep{nm512_dreamerv3-torch_2023} and on the Pytorch implementation of the S5 layer \citep{c2d_s5_2023}. As this version does not implement the resettable version of S5 and \citet{lu_structured_2023} do not provide an open source implementation of their method, we implemented the reset mechanism ourselves in the provided source code. The source code is publicly available under the following URL: \repourl 

\section{Ablations}
\label{sec:appC}


In the following, we provide additional ablation studies exploring the effect of different components of \hieros. We conduct all ablations on four games: 
(i) Krull, a game that features multiple levels, 
(ii) Breakout, a game with a single level and simple dynamics, 
(iii) Battle Zone, a game with a single level and complex hierarchical dynamics and 
(iv) Freeway, a game with difficult exploration properties \citep{micheli_transformers_2023}. We use the same hyperparameters as described in \Cref{sec:appB} for all ablations. We show the collected reward of the lowest level subactor of \hieros with S5WM and all hyperparameters as specified in \Cref{sec:appB} in red and the collected reward of \hieros with the ablation in blue. We use the same color scheme for all ablation studies, except those where the graphs contain more than two lines. We conducted three training runs for each ablation, and the graphs illustrate the average performance across those runs as a line. The shaded area around the line, in the same color, represents the range of values observed during the three runs. In every interaction with the environment, the actor's action is repeated for four frames, aligning with the methodology adopted in other papers such as \citet{hafner_mastering_2022, hafner_mastering_2023, micheli_transformers_2023, robine_transformer-based_2023}. Consequently, the plots display results up to 400k steps, but it's important to note that only 100k interactions were actually conducted in the specified environments due to the frame repetition.

\subsection{S5WM vs. RSSM}
\label{ablation:rssm}

\begin{figure*}[ht]  
    \centering
    \includegraphics[width=0.8\textwidth]{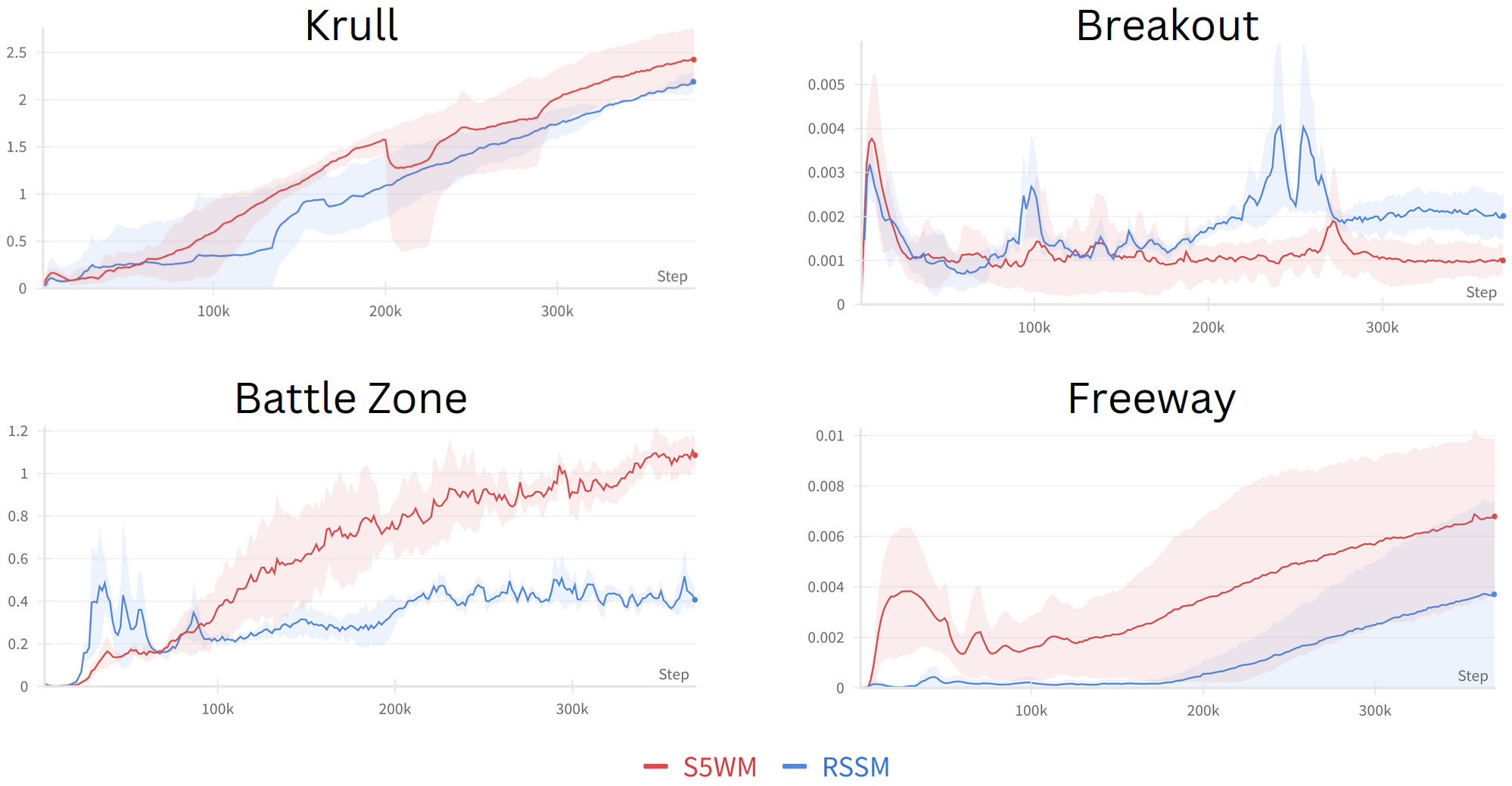}
    \caption{Collected rewards per step for Krull, Breakout, Battle Zone, and Freeway of the lowest level subactor for \hieros with an RSSM (blue) and \hieros with an S5WM (red).}
\label{fig:rssm}
\end{figure*}

In \Cref{sec:imagined_trajectories}, we compare the model losses and partial rewards of \hieros in the game of Krull and Breakout using either RSSMs or S5WMs as world models. In \Cref{fig:rssm}, we compare the collected rewards of \hieros using either RSSMs (blue) or S5WMs (red) for Krull, Breakout, Battle Zone, and Freeway against each other. The RSSM is not able to achieve the same performance as the S5WM for Krull, Battle Zone and Freeway. However, for Breakout, the RSSM exhibits a slight improvement compared to the S5WM. This aligns with the observation made in Section \ref{sec:imagined_trajectories}, where it was noted that the world model losses for Breakout were lower when employing an RSSM as the dynamics model compared to using the S5WM.

\subsection{Internal S5 Layer State as Deterministic World State}
\label{sec:appC:state_as_deter}

\begin{figure*}[ht]
    \centering
    \includegraphics[width=0.8\textwidth]{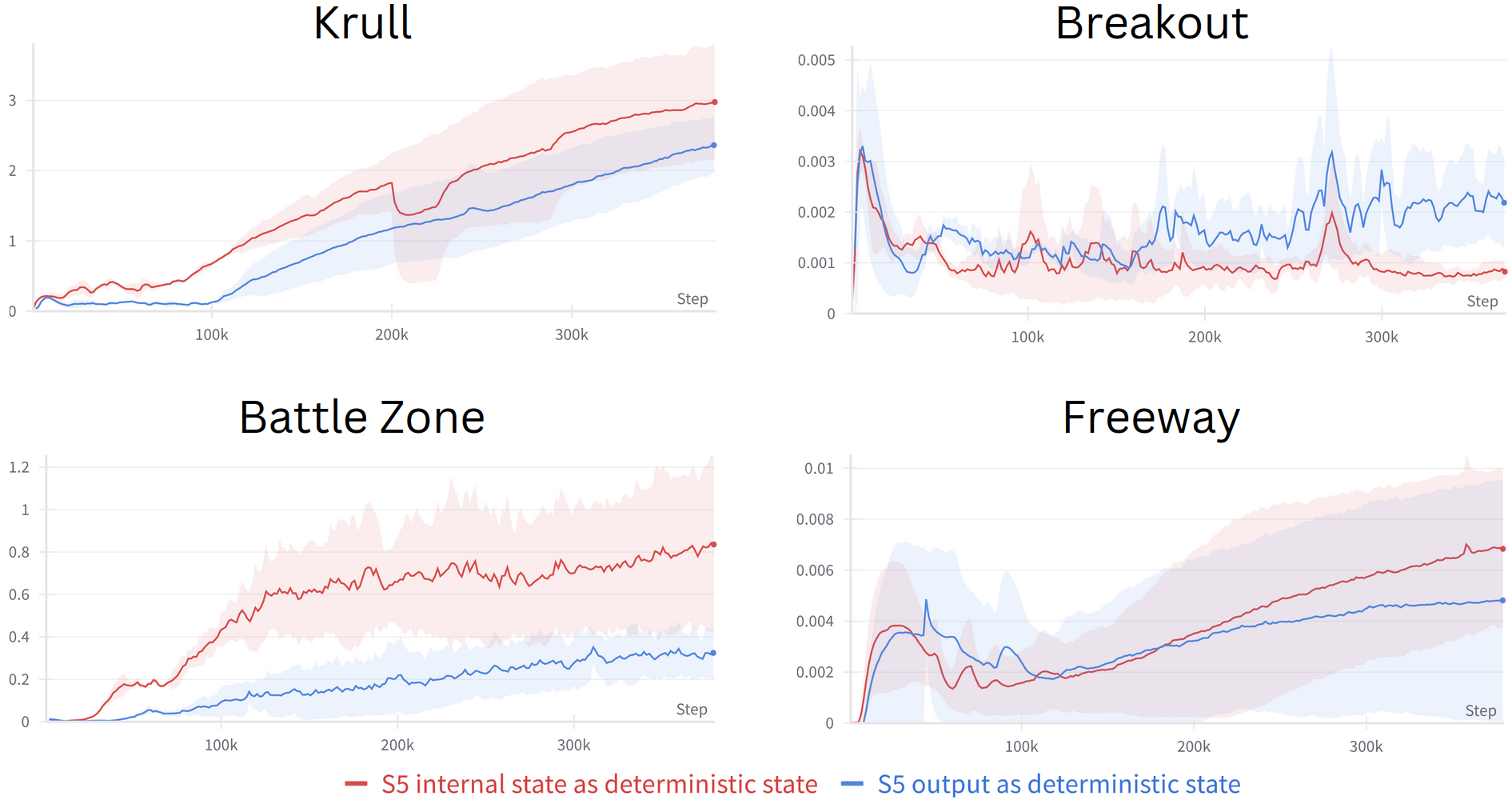}
    \caption{Comparison of the collected reward per step of \hieros with S5WM using the internal state $x_t$ of the stacked S5 layers as the deterministic part of the latent state $h_t$ (red) and using the output of the stacked S5 layers as $h_t$ (blue) for Krull, Breakout, Battle Zone, and Freeway.}
    \label{fig:state_deter}
\end{figure*}

In \Cref{sec:s5wm}, we describe the S5WM, which uses the S5 layer to predict the next world state.
In this section, we explore the effect of using the internal state $x_t$ of the stacked S5 layers of S5WM as the deterministic part of the latent state $h_t$ instead of the output of the S5 layers. Other comparable approaches that swap out the GRU in the RSSM with a sequence model usually use the output of the sequence model as $h_t$  \citep{chen_transdreamer_2022,micheli_transformers_2023,robine_transformer-based_2023,deng_facing_2023}. So, testing this ablation provides valuable insight into how the learned world states can be enhanced in order to boost prediction performance. \Cref{fig:state_deter} shows the influence of using the internal state as $h_t$ vs. using the output of the S5 layers as $h_t$ for \hieros. Using the internal S5 layer state as deterministic world model state seems to boost the performance for Krull and Battle Zone while having no clear benefit for Breakout and Freeway.

\subsection{Hierarchy Depth}
\label{sec:appC:model_hierarchy_depth}

\begin{figure*}[ht]
    \centering
    \includegraphics[width=0.8\textwidth]{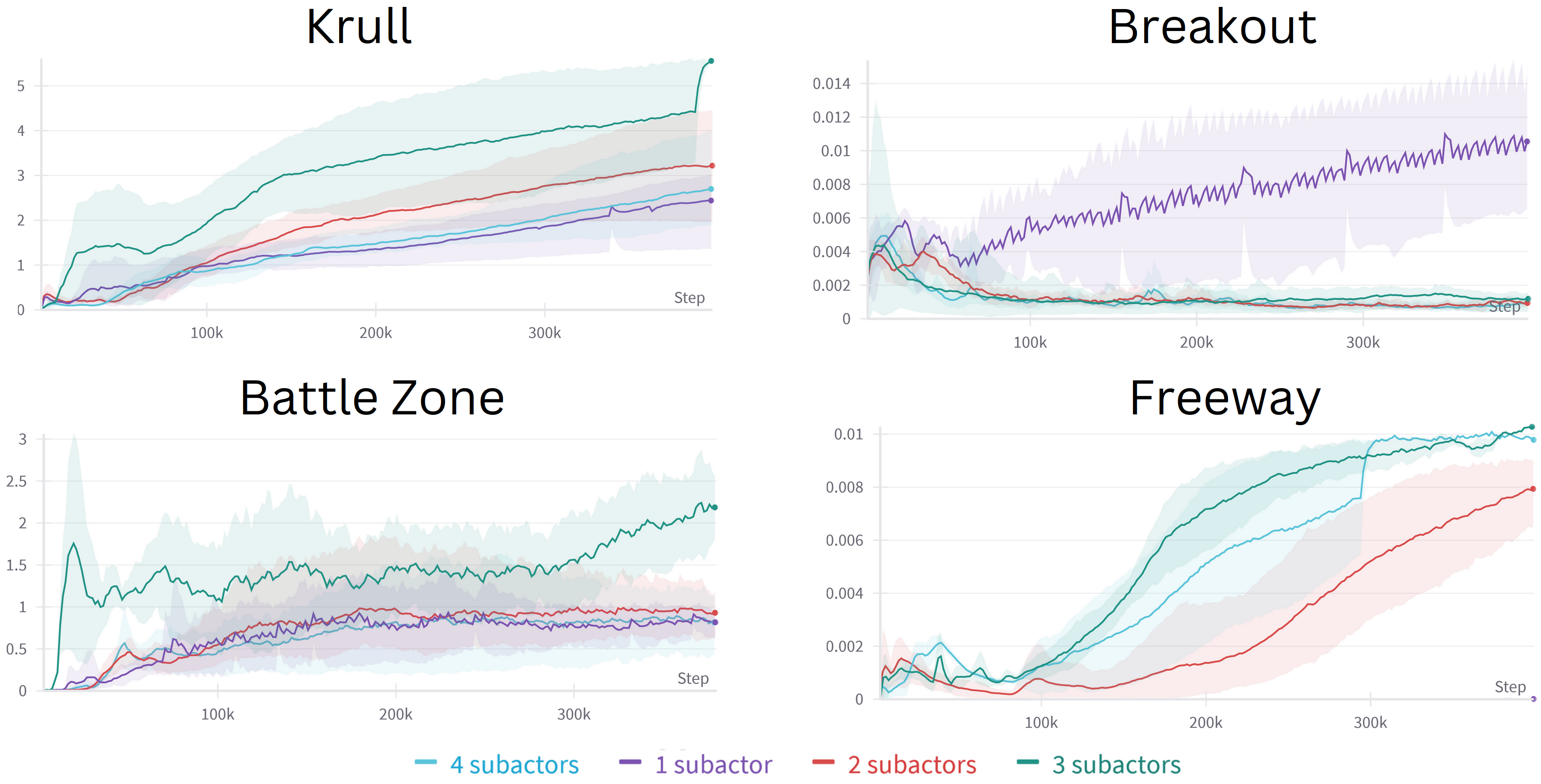}
    \caption{Comparison of the collected reward per step of \hieros with S5WM using a model hierarchy depth of 1 (purple), 2 (red), 3 (green), and 4 (light blue) for Krull, Breakout, Battle Zone, and Freeway.}
    \label{fig:num_layers}
\end{figure*}

In this section, we show the effect of using different model hierarchy depths. We compare \hieros with S5WM using a model hierarchy depth of 1, 2, 3, and 4. \Cref{fig:num_layers} shows the collected reward of \hieros with S5WM using a model hierarchy depth of 1 (purple), 2 (red), 3 (green), and 4 (light blue) for Krull, Breakout, Battle Zone, and Freeway. Most remarkably, we can see that \hieros with just one layer achieves significantly better results than with two or more subactors. This indicates that a single layer algorithms, like DreamerV3, Iris, or TWM have a significant advantage over multi-layer algorithms, like \hieros in games with no distribution shifts and easy to predict dynamics.

\subsection{Uniform vs. Time-Balanced Replay Sampling}
\label{sec:appC:uniform_vs_timebalanced_replay_sampling}

\begin{figure*}[ht]
    \centering
    \includegraphics[width=0.8\textwidth]{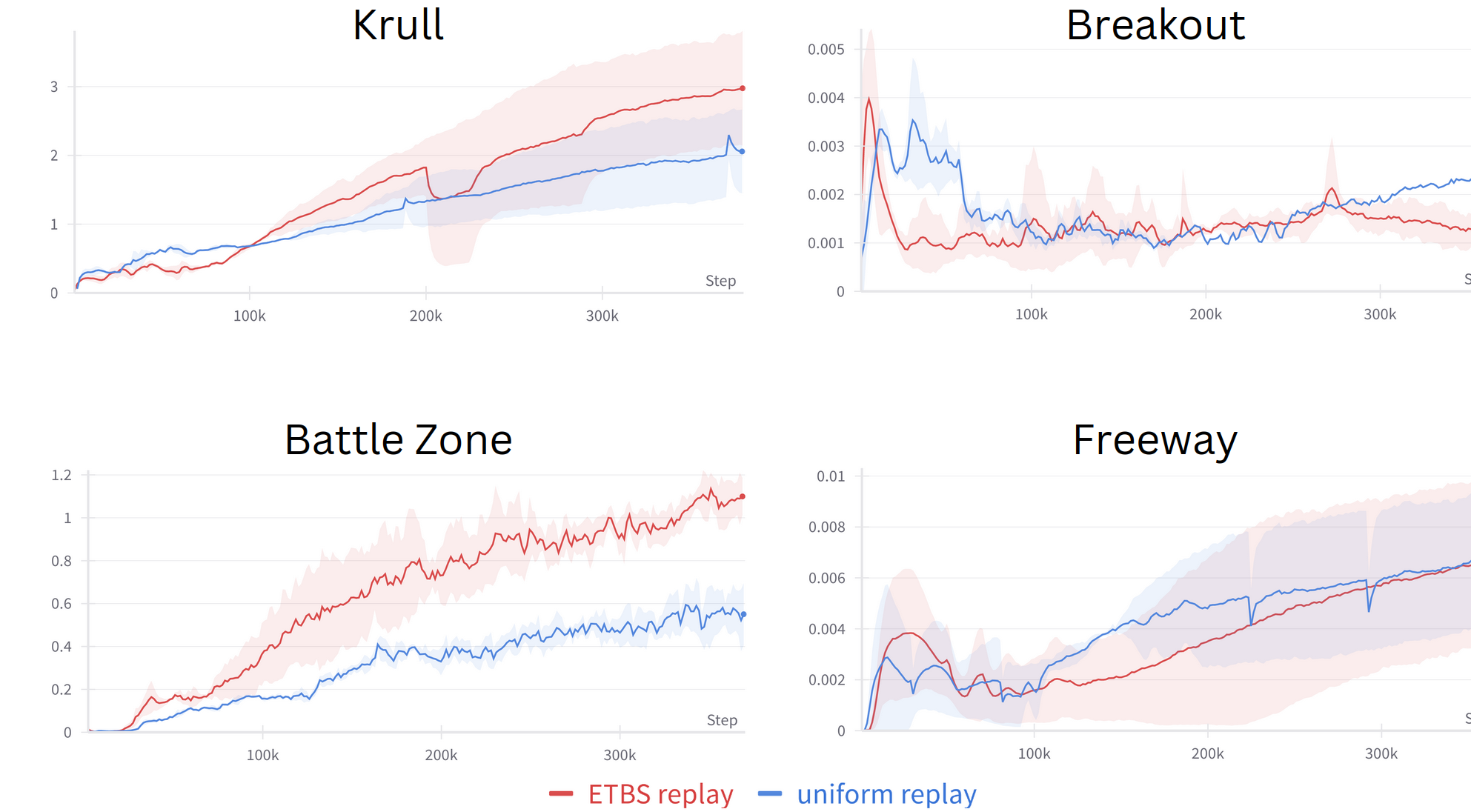}
    \caption{Comparison of the collected reward per step of \hieros with S5WM using uniform sampling (red) and time-balanced sampling (blue) for Krull, Breakout, Battle Zone, and Freeway.}
    \label{fig:uniform}
\end{figure*}

In \Cref{sec:efficient_time_balanced_sampling}, we describe the efficient time-balanced sampling method. In this section, we compare the effect of using uniform sampling vs. our efficient time-balanced sampling for the experience dataset. \Cref{fig:uniform} shows the collected reward of \hieros with S5WM using uniform sampling (red) and time-balanced sampling (blue) for Krull, Breakout, Battle Zone, and Freeway. As can be seen, in the case of Freeway and Breakout, the time balanced sampling did not provide a significant advantage, while it boosted the performance considerably for Krull and Battle Zone, two games that both display hierarchical challenges. This indicates, that in cases where the model hierarchy can contribute a lot to the overall performance, the actor is more sensitive to overfitting on older training data, containing subgoals produced by less trained higher level subactors.

\subsection{Providing k World States vs. Only the k-th World State as Input for the Higher Level World Model}
\label{sec:appC:intermediate_encoding_vs_only_kth}

\begin{figure*}[ht]
    \centering
    \includegraphics[width=0.8\textwidth]{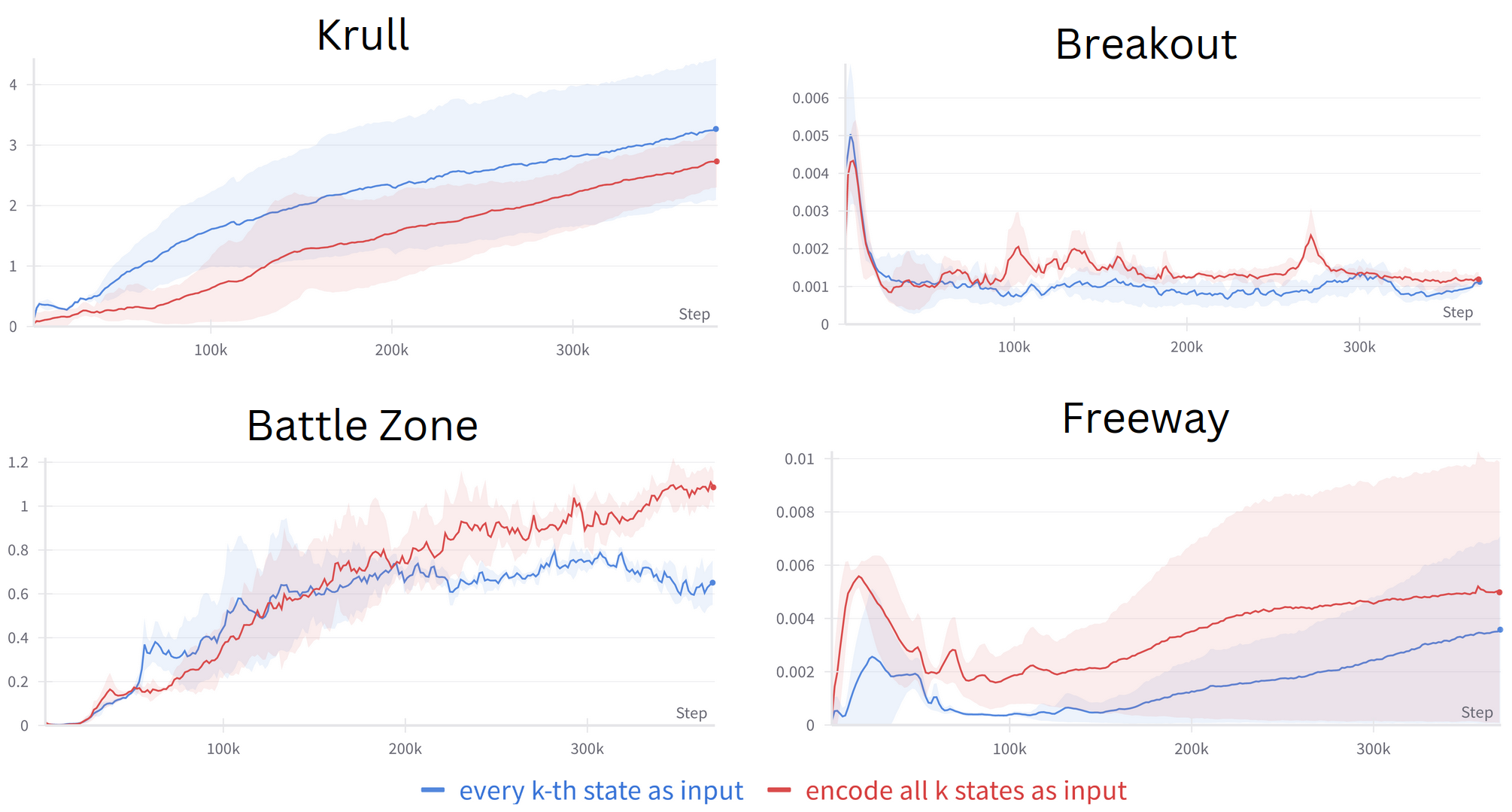}
    \caption{Comparison of the collected reward per step of \hieros with S5WM using $k$ consecutive world states as input for the higher level world model (red) and only the $k$-th world state as input for the higher level world model (blue) for Krull, Breakout, Battle Zone and Freeway.}
    \label{fig:intermediate_encoding_vs_only_kth}
\end{figure*}

In \Cref{sec:multilayered_hierarchical_imagination}, we describe how \hieros provides $k$ consecutive world states of the lower level world model as input for the higher level subactor. However, many approaches such as Director \citep{hafner_deep_2022} only provide the $k$-th world state as input for the higher level world model. In this section, we explore the effect of providing only the $k$-th world state as input for the higher level world model. \Cref{fig:intermediate_encoding_vs_only_kth} shows the collected reward of \hieros with S5WM using $k$ consecutive world states as input for the higher level world model (red) and only the $k$-th world state as input for the higher level world model (blue) for Krull, Breakout, Battle Zone, and Freeway. Providing only the $k$-th world state as input seems to deteriorate the performance for Battle Zone and Freeway while being beneficial for Krull.

\subsection{Using a Smaller Version of S5WM}
\label{app:smol_s5}

\begin{figure*}[ht]
    \centering
    \includegraphics[width=0.8\textwidth]{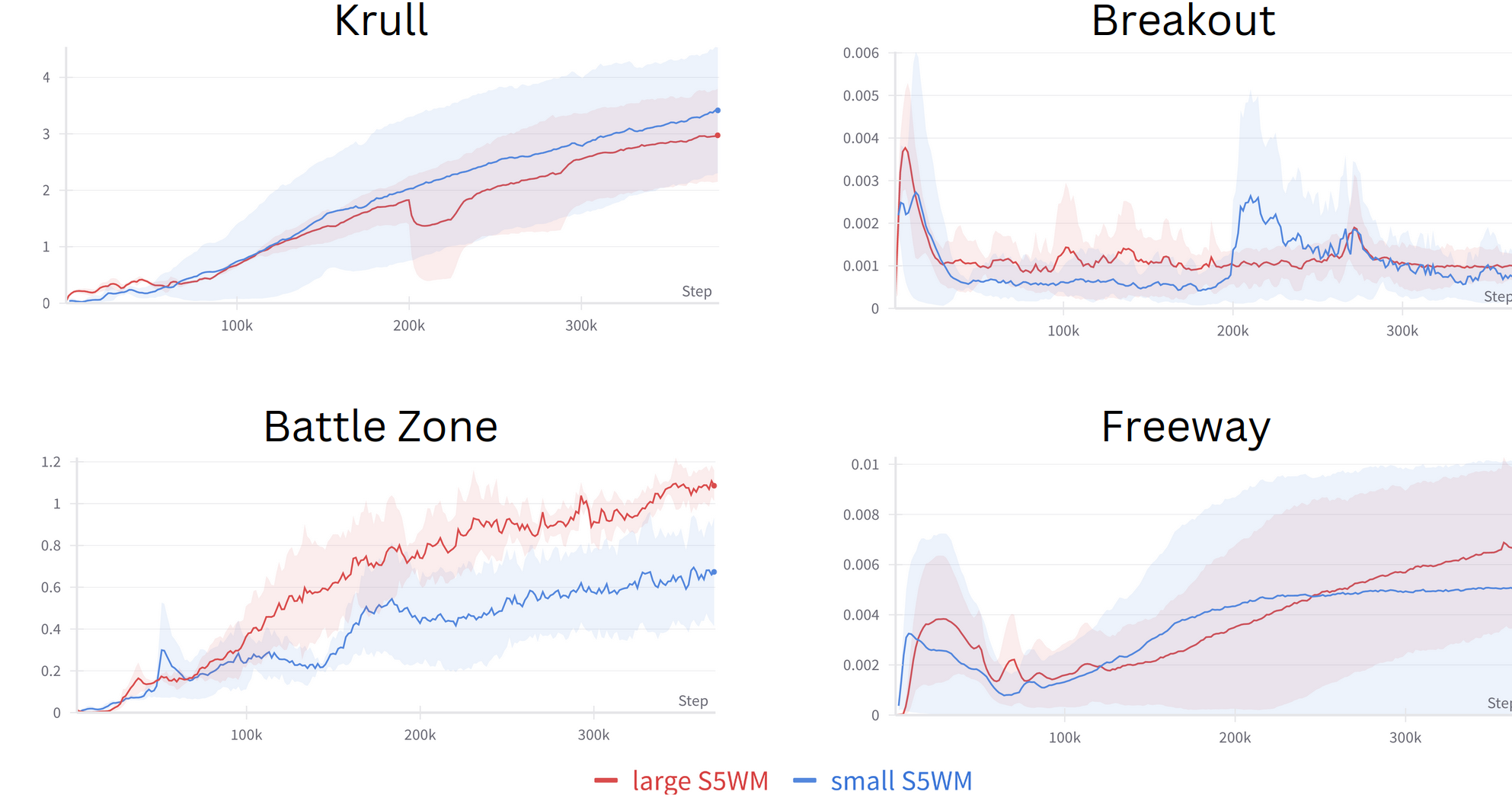}
    \caption{Comparison of the collected reward per step of \hieros with the standard S5WM (red) and \hieros using a smaller version of S5WM (blue) for Krull, Breakout, Battle Zone and Freeway.}
    \label{fig:smol_s5wm}
\end{figure*}

In \cref{sec:imagined_trajectories}, we propose a theory that suggests larger S5WM models yield superior results in complex environments, whereas smaller dynamic models may be more advantageous in simpler dynamic environments such as Breakout. In \cref{fig:smol_s5wm}, we compared \hieros using the standard-sized S5WM (in red) with \hieros using a smaller S5WM, which consists of only 4 S5 blocks and is thus half the size (in blue). In the case of Battle Zone, a game characterized by complex dynamics, the larger S5WM significantly enhances performance. Conversely, for Breakout, we observe that the smaller S5WM performs better during certain stages of training but exhibits a decline towards the end. This suggests that the smaller S5WM could potentially outperform the larger counterpart in this environment, provided additional measures are implemented to prevent actor degeneration during later training stages.

\subsection{Providing Decoded Subgoals as Actor Input}
\label{app:decompress}

\begin{figure*}[ht]
    \centering
    \includegraphics[width=0.8\textwidth]{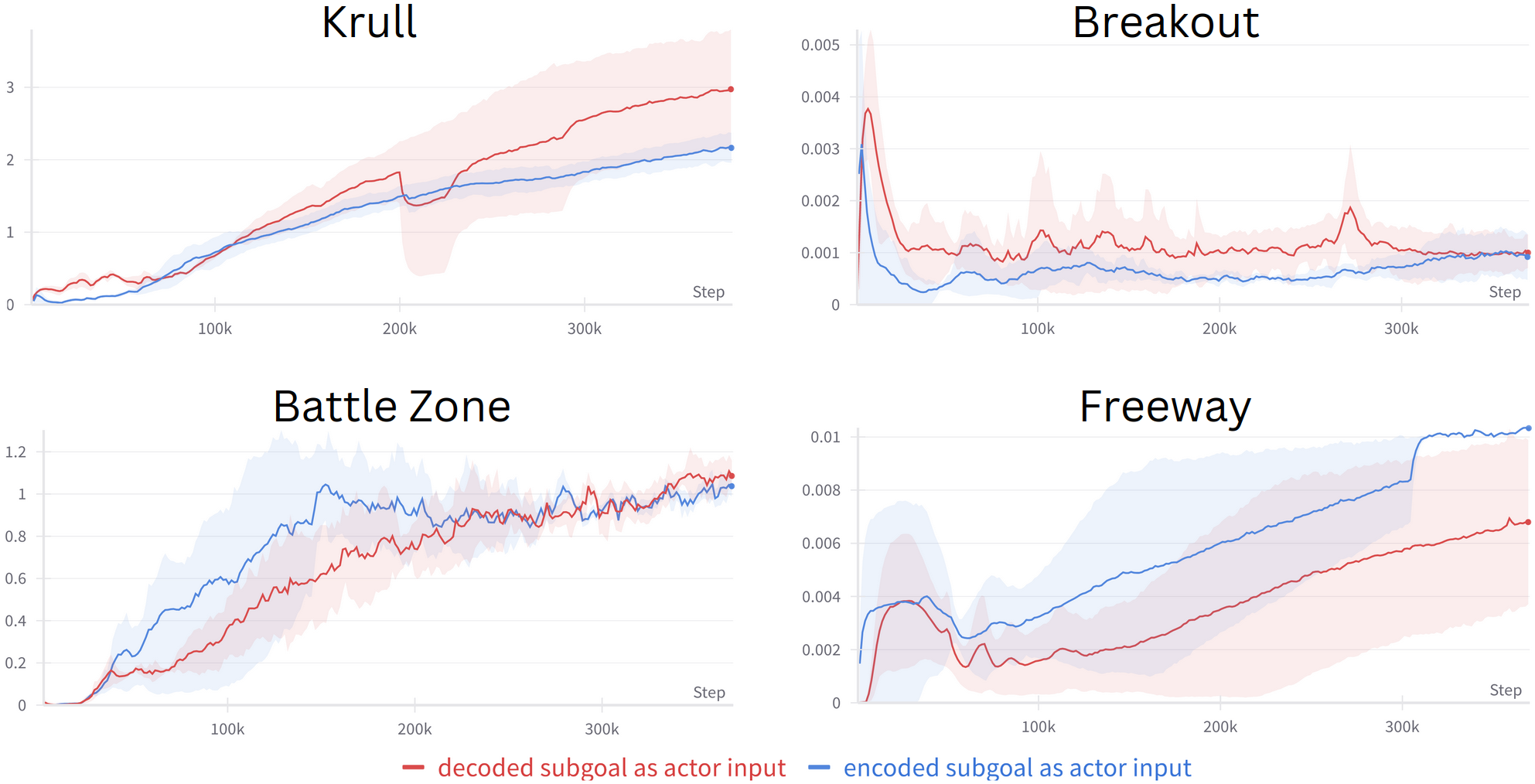}
    \caption{Comparison of the collected reward per step of \hieros when passing encoded (red) and when passing subgoals decoded by the subgoal autoencoder (blue) to the lower level subactors for Krull, Breakout, Battle Zone and Freeway.}
    \label{fig:decompress}
\end{figure*}

In \cref{sec:multilayered_hierarchical_imagination}, we explain that in \hieros, the encoded subgoal $g^i$ from the higher-level policy is passed as input to the lower-level input policy. This differs from the approach in \citet{hafner_deep_2022}, where decoded subgoals from the subgoal autoencoder are passed to the worker policy in their hierarchical model. In \cref{fig:decompress}, we directly compare the impact of using encoded (in red) and decoded (in blue) subgoals as input for the lower-level subactor. Notably, utilizing encoded subgoals appears to enhance performance in Krull and Breakout but leads to a significant degradation in performance for Freeway. In practical terms, a tradeoff must be considered based on the specific environment. It's worth noting that passing decompressed subgoals increases the overall parameter count of the model, but in cases like Freeway, the additional model size contributes to further improvements in rewards.

\subsection{Further Ablations Left for Future Work}

There are multiple other interesting directions for further ablation studies: One of the main novelties of \hieros is its use of hierarchical world models. Architectures like Dreamer \citep{hafner_deep_2022} however train both the higher and the lower level policy on the same world model, so exploring this might give valuable insights, how much the hierarchical world models contribute to the performance of \hieros. Another interesting direction is to explore the effect of using differently sized subactors, with the lowest level subactor having the largest amount of trainable parameters and the higher levels having less and less trainable parameters. This could potentially lead to a more efficient use of the available parameters. The higher levels are trained fewer times and with fewer data than the lower levels, so smaller networks might speed up training in those cases. 

For \hieros we relied on the proven world model architecture used by the DreamerV3 model, which features a world state composed of a deterministic and a stochastic part. However, other approaches like IRIS \citep{micheli_transformers_2023} do not rely on this stochastic world state and achieve comparable result. So exploring the effect of using only a deterministic world state might be interesting.

\section{CDF of Imbalanced Replay Sampling}
\label{sec:appD:sampling}

In \Cref{sec:efficient_time_balanced_sampling}, we use the CDF of the skewed sampling distribution that arises when iteratively adding elements to a dataset and sampling uniformly from it. The distribution is defined as follows:
\begin{align}
    p(x) = \frac{H_n - H_x}{n} \approx \frac{\ln(n) - \ln(x)}{n} = \tilde p(x)
\end{align}
with $n$ being the size of the dataset. Since the inequality $\ln(x) < H_x < 1 + \ln(x)$ holds for all $x \geq 2$ we assume without loss of generality that $2 \leq x \leq n$. However, since $\tilde p(x)$ does not sum up to 1 over the interval $\left[2, n\right]$, we need to divide it by its integral:
\begin{align}
    p_{s}(x) &= \frac{\tilde p(x)}{\int_2^n \tilde p(x) dx} 
    = \dfrac{\frac{\ln(n) - \ln(x)}{n}}{\frac{-2\ln\left(n\right)+n+2\ln\left(2\right)-2}{n}} \\
    &= \dfrac{\ln(n) - \ln(x)}{-2\ln\left(n\right)+n+2\ln\left(2\right)-2} \nonumber
\end{align}
with $p_s$ being the approximate probability density function of the skewed sampling distribution.
The CDF of this distribution is defined as follows:
\begin{align}
    CDF_{s}(x) = \int_2^x p_{s}(x) dx 
    = P_{s}(x) - P_{s}(2)
\end{align}
with $P_{s}(x)$ being the antiderivative of $p_{s}(x)$:
\begin{align}
    P_{s}(x) &= \int p_{s}(x) dx
    = \dfrac{x\cdot\left(\ln\left(x\right)-\ln\left(n\right)-1\right)}{2\ln\left(n\right)-n-2\ln\left(2\right)+2}
\end{align}

\begin{figure*}[ht]
    \centering
    \includegraphics[width=0.8\textwidth]{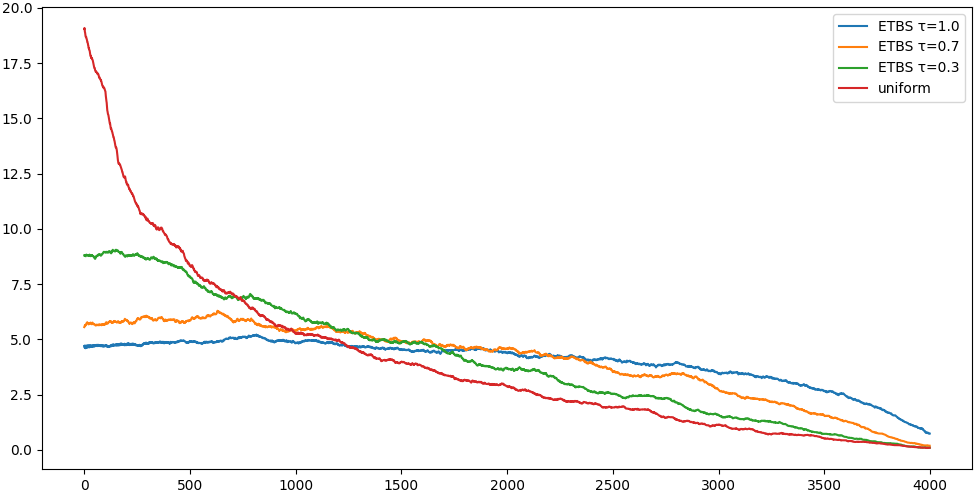}
    \caption{The number of times an entry $i$ is sampled in a dataset of final size 4000 with iteratively adding one element to the dataset and sampling over the dataset afterwards. Compared are the original uniform sampling method implemented in DreamerV3 \citep{hafner_mastering_2023} and our proposed ETBS with different temperatures $\tau$. For our final experiments, we use a temperature of $\tau = 0.3$.}
    \label{fig:samplings}
\end{figure*}

With this, we can derive $CDF_{s}(x)$ in closed form:
\begin{align}
    CDF_{s}(x) = \dfrac{x\cdot\left(\ln\left(x\right)-\ln\left(n\right)-1\right)+2\left(\ln\left(n\right)-\ln\left(2\right)+1\right)}{2\ln\left(n\right)-n-2\ln\left(2\right)+2}
\end{align}
This can be computed in $O(1)$ time and is therefore very efficient. With the $CDF_{s}(x)$ we can calculate the efficient time-balanced sampling distribution $p_{etbs}(x)$ as described in \Cref{sec:efficient_time_balanced_sampling}. It is also important to mention that we assume, that we only add one item to the dataset and then sample one time after each step. However, if we add $k$ items before sampling $s$ times, the expected number of draws for one element becomes $\mathbb{E}({N_{x_i}}) = \frac{k \cdot (H_n - H_i)}{s \cdot n}$. This additional factor $\frac{k}{s}$ is canceled out when computing the probabilities for one element from the expected value, which is why we can ignore this in our computations. 
\Cref{fig:samplings} shows the sampling counts for different temperatures $\tau$ for ETBS.

\section{Further Future Work}
\label{sec:future_work}

Another possible future direction would be to use the S5-based actor network proposed by \citet{lu_structured_2023} for \hieros. Right now, the imagination procedure still relies on a single step prediction of the world model due to the single step architecture of the actor network. Using the S5-based actor network would allow for a multistep imagination procedure, which could further improve the performance of \hieros. This would also open the door to more efficient look-ahead search methods.

Since S4/S5-based models and Transformer-based models show complementary strengths in regard to short and long-term memory recall, many hybrid models have been proposed \citep{dao2022hungry, zuo2022efficient, mehta2022long, gupta2022simplifying}. Exploring these more universal models might also be a promising direction for future research.

\citet{lecun2022path} describes a modular RL architecture, which combines HRL, world models, intrinsic motivation, and look-ahead planning in imagination as a potential candidate architecture for a true general intelligent agent. They propose learning a reconstruction free latent space to prevent a collapse of the learned representations, which is already explored in several works for RL \citep{okada_dreaming_2021,schwarzer_pretraining_2021}. They also describe two modes of environment interaction: reactive (Mode 1) and using look-ahead search (Mode 2). \hieros implements several parts of this architecture, namely the hierarchical structure, hierarchical world models and the intrinsic motivation. \hieros, like most RL approaches, uses the reactive mode, while approaches like EfficientZero \citep{ye2021mastering} could be interpreted as Mode 2 actors. \citet{koul2020dream} implement a Monte Carlo Tree Search (MCTS) planning method in the imagination of a Dreamer world model. Implementing similar methods for the hierarchical structure of \hieros, combined with the more efficient S5-based world model (potentially also an S5 based actor network) could yield a highly efficient planning agent capable of learning complex behavior in very dynamic and stochastic environments.

Since \Cref{fig:num_layers} demonstrates that for some environments a deeper hierarchy can deteriorate performance, a possible future research could include an automatic scaling of the hierarchy depending on the current environment. E.g. if the accumulated reward stagnates and the agent cannot find a policy that further improves performance, the agent might automatically add another hierarchy layer, perform a model parameter reset and retrain on the collected experience dataset.

\end{document}